\documentclass[sn-mathphys]{sn-jnl}

\jyear{2023}%

\theoremstyle{thmstyleone}%

\theoremstyle{thmstyletwo}%

\theoremstyle{thmstylethree}%

\usepackage{caption}
\usepackage{subcaption}
\usepackage[export]{adjustbox}
\usepackage{amsmath}
\usepackage{amssymb}
\usepackage{multirow}
\usepackage{mathtools}
\usepackage{booktabs}
\usepackage{tabularx}
\usepackage{graphicx}
\usepackage{booktabs}

\usepackage{breqn}

\usepackage{float}
\usepackage{lipsum}
\usepackage{ifmtarg}
\usepackage{array}

\usepackage[utf8]{inputenc}
\usepackage{geometry}

\begin{document}

\title[Class Imbalanced Learning with SVM]{Methods for Class-Imbalanced Learning with Support Vector Machines: A Review and an Empirical Evaluation}

\author[1]{\fnm{Salim} \sur{Rezvani}}\email{salim.rezvani@torontom.ca}

\author*[2]{\fnm{Farhad} \sur{Pourpanah}}\email{farhad.086@gmail.com }

\author[3]{\fnm{Chee Peng} \sur{Lim}}\email{chee.lim@deakin.edu.au}

\author[3]{\fnm{Q. M. Jonathan} \sur{Wu}}\email{jwu@uwindsor.ca}

\affil[1]{\orgdiv{Department of Computer Science,} \orgname{Toronto Metropolitan University,} \orgaddress{\city{Toronto,} \state{Ontario,} \country{Canada}}}

\affil[2]{\orgdiv{Department of Electrical and Computer Engineering,} \orgname{Queen's University,} \orgaddress{ \city{Kingston,} \state{Ontario,} \country{Canada}}}

\affil[3]{\orgdiv{Institute for Intelligent Systems Research and Innovation}, \orgname{Deakin University,} \orgaddress{ \city{Geelong,} \state{Victoria,} \country{Australia}}}

\affil[4]{\orgdiv{Department of Electrical and Computer Engineering,} \orgname{University of Windsor,} \orgaddress{ \city{Windsor,} \state{Ontario,} \country{Canada}}}

\abstract{ This paper presents a review on methods for class-imbalanced learning with the Support Vector Machine (SVM) and its variants. We first explain the structure of SVM and its variants and discuss their inefficiency in learning with class-imbalanced data sets. We introduce a hierarchical categorization of SVM-based models with respect to class-imbalanced learning. Specifically, we categorize SVM-based models into re-sampling, algorithmic, and fusion methods, and discuss the principles of the representative models in each category. In addition, we conduct a series of empirical evaluations to compare the performances of various representative SVM-based models in each category using benchmark imbalanced data sets, ranging from low to high imbalanced ratios. Our findings reveal that while algorithmic methods are less time-consuming owing to no data pre-processing requirements, fusion methods, which combine both re-sampling and algorithmic approaches, generally perform the best, but with a higher computational load. A discussion on research gaps and future research directions is provided.

}


\maketitle

\section{Introduction}
\label{Sec:intro}
Learning class imbalanced data, i.e., one or more target (majority) classes contain far more samples than those from other (minority) classes, poses a major challenge in supervised learning. In many real-world applications, such as medical diagnosis~\cite{wang2021fuzzy}, credit card fraud detection~\cite{randhawa2018credit}, condition monitoring systems~\cite{pourpanah2018anomaly} and natural disasters~\cite{maalouf2011robust}, it is difficult and expensive to obtain data samples with respect to minority (e.g., abnormal) or minority classes~\cite{pourpanah2020review,he2009learning,wang2020recent}. As such, the number of samples belonging to the majority or normal classes is often larger than that of the minority classes. Standard supervised classification algorithms, which work well with  less skewed data sample distribution problems, can become biased towards the majority classes; therefore failing to yield favorable accuracy scores for the minority classes~\cite{castano2020enhanced}.\par

Over the years, many methods have been developed to enhance the performance of standard classification algorithms in learning class-imbalanced data.
In general, the available methods to solve this issue can be categorized into three groups: re-sampling, algorithmic, and fusion methods~\cite{liu2021fuzzy}. Re-sampling methods, which focus on data pre-processing, change the prior distributions of both minority and majority classes~\cite{barua2012mwmote}. Algorithmic methods modify the algorithm structure to enhance their robustness in learning imbalanced data~\cite{khan2017cost}. Fusion methods combine different techniques or ensemble models for addressing class imbalanced data~\cite{sun2015novel}.\par

The support vector machine (SVM)~\cite{cortes1995support,Vapnik2000the} is an effective kernel-based learning algorithm for solving data classification problems. It aims to find a hyperplane to optimally separate data samples with maximal margins~\cite{tao2013hessian,rezvani2019intuitionistic,cristianini2000an}. Due to its structural risk minimization capability, SVM can effectively reduce overfitting and avoid local minima~\cite{kang2018adistance}. Since its introduction, many SVM variants have been proposed, e.g., fuzzy SVM (FSVM)~\cite{lin2002fuzzy}, twin SVM (TSVM)~\cite{khemchandani2007twin}, fuzzy TSVM (FTSVM)~\cite{gao2015coordinate}, intuitionistic FTSVM (IFTSM)~\cite{rezvani2019intuitionistic}, etc. Despite their success in handling balanced data sets, their performances are severely affected by imbalanced data distribution~\cite{koknar2009improving}. This is caused by treating all samples equally and ignoring the difference among both majority and minority classes.  As a result, a decision boundary biases toward the majority class samples is established when the underlying SVM models learn from imbalanced data.\par

There are several reviews of imbalanced classification methods in the literature~\cite{he2009learning,akbani2004applying,galar2012areview,haixiang2017learning,he2013class}. In 2004, the study in~\cite{akbani2004applying} discussed the weaknesses of traditional SVM models in solving imbalanced classification problems and explained why the conventional under-sampling methods are not the best choice. In 2009, the study in~\cite{he2009learning} provided a critical review of imbalanced learning methods along with several assessment metrics. Moreover, challenging issues and future research directions were presented.
In 2012, the study in~\cite{galar2012areview} introduced a taxonomy of ensemble-based methods to address class imbalanced issues and provided empirical evaluations with respect to the representative ones of each category. On the other hand, the study in~\cite{haixiang2017learning} presented an in-depth review of rare event detection in imbalanced learning problems between 2006 and 2016. In 2013, the study in~\cite{batuwita2013class} reviewed SVM-based re-sampling and algorithmic methods. While these surveys have been conducted years ago, none of them provide an in-depth and comprehensive review of SVM-based methods for addressing class-imbalanced learning problems. As such, our review presented in this paper aims to fill this gap.  
Firstly, we explain SVM and its variants and discuss why they cannot effectively learn from class imbalanced data. Then, we provide a hierarchical categorization of SVM-related methods for tackling class imbalanced problems and present the representative ones of each category. In addition, we conduct a thorough empirical comparison of these methods. In short, the main contributions of this paper are as follows:
\begin{itemize}
\item  a comprehensive review of SVM-based methods for tackling class imbalanced problems;
\item a hierarchical categorization of imbalanced SVM methods along with the representative ones of each category; 
\item a comprehensive performance evaluation of SVM-based class imbalanced learning methods;
\item a discussion on the key research gaps and future research directions.
\end{itemize}

This review paper consists of five sections. Section~\ref{Sec:pre} explains the structures of SVM, FSVM and TSVM, and discusses why these methods are not able to handle imbalanced classification problems. Section~\ref{Sec:isvm} reviews imbalanced SVM-based methods. Specifically, a hierarchical categorization of  imbalanced SVM-based methods is presented, and each category is further divided into several constituents. Section~\ref{Sec:emp} presents an empirical study on the performance of various SVM-based models. The results are discussed and analyzed. Finally, Section~\ref{Sec:dis} presents the concluding remarks and suggestions for future research. 
\section{Preliminaries}
\label{Sec:pre}
In this section, we explain the structures and principles of SVM, FSVM and TSVM.  Importantly, we discuss the issues of these models in handling imbalanced data classification problems. 

\vspace{-0.2cm}
\subsection{Support Vector Machine}
\label{Sec:sec:svm}
Let $\{(\textbf{x}_i,y_i)\}_{i=1}^N$ represent a binary classification problem where $\textbf{x}_i\in R^D$ is a $D$-dimensional input data sample and $y_i\in\{-1, 1\}$ indicates the target class. To learn an optimal hyperplane, SVM transforms the input sample into a high dimensional feature space using a non-linear mapping function $\Phi$. The optimal hyperplane pertaining to the transformed feature space can be written as follows:
\begin{align}
    \label{eq:svm1}
    w.\Phi(x)+b=0
\end{align}
where $w\in R^D$ and $b\in R$ are the weight vector and bias, respectively. For linear classification problems, the maximal margin optimization problem is solved to obtain the hyperplane, as follows:
\begin{align}
    \label{eq:svm_lin}
   \min(\frac{1}{2}w.w),
\end{align}
\begin{align*}
 \textit{s.t.}~~y_{i}(w.\Phi(x_{i})+b)\geq1,~~i=1,2,...,N   
\end{align*}

For non-linear classification problems, even though the input samples are transformed into a high-dimensional feature space, they are not completely separable by linear functions. To alleviate this issue, a set of slack variables $\xi_i\geq0$ is integrated into (\ref{eq:svm_lin}): 
\begin{align}
    \label{eq:svm_non}
    \min(\frac{1}{2}w.w+C\sum_{i=1}^{N}\xi_{i})
\end{align} 
\begin{align*}
\textit{s.t.} & y_{i}(w.\Phi(x_{i})+b)\geq1-\xi_{i}
\end{align*}
\begin{align*}
\xi_{i}\geq0,~~i=1,2,...,N.
\end{align*}
where $\sum_{i=1}^N\xi_i$ is the penalty factor that measures the total error, and $C$ is a parameter to balance the trade-off between maximizing the margin and minimizing the error. To solve this quadratic optimization problem, the Lagrangian optimization is constructed, as follows:
\begin{align}
\label{Eq:svm_lag}
\max\{\sum^{l}_{i=1}\alpha_{i}-\frac{1}{2}\sum^{l}_{i=1}\sum^{l}_{j=1}\alpha_{i}\alpha_{j}y_{i}y_{j}\Phi(x_{i})\Phi(x_{j})\}
\end{align}
\begin{align*}
\textit{s.t.}~\sum^{l}_{i=1}y_{i}\alpha_{i}=0,~0\leq\alpha_{i}\leq C,~i=1,2,...,N,
\end{align*}
where $\alpha_{i}$~$(i=1,2,...,l)$ represent the Lagrangian multipliers that satisfy the Karush-Kuhn-Tucker (KKT)~\cite{yu2007novel} conditions:
\begin{align}\label{Eq:kkt1}
\alpha_i (y_i(w.\phi(x_i)+b)-1+\xi_i)=0, ~i=1,2,...,N,
\end{align}
\begin{align}\label{Eq:kkt2}
(C-\alpha_{i})\xi_{i}=0,~i=1,2,...,N.
\end{align}

In SVM, it is not necessary to know the exact mapping function $\Phi(x_{i})$. As such, a kernel function, i.e., $K(x_{i},x_{j})=\Phi(x_{i}).\Phi(x_{j})$, can be applied to transform the dual optimization problem in (\ref{Eq:svm_lag}) into:
\begin{align}\label{Eq:ker}
max\{\sum^{N}_{i=1}\alpha_{i}-\frac{1}{2}\sum^{N}_{i=1}\sum^{N}_{j=1}\alpha_{i}\alpha_{j}y_{i}y_{j}K(x_{i},x_{j})\}
\end{align}
\begin{align*}
s.t.~\sum^{l}_{i=1}y_{i}\alpha_{i}=0,~0\leq\alpha_{i}\leq C,~i=1,2,...,N
\end{align*}

{ After finding the optimal values of $\alpha_i$ by solving~(\ref{Eq:ker}), $w$ is obtained as follows:}
\begin{align}\label{eq:www}
w=\sum^{N}_{i=1}\alpha_{i}y_{i}\Phi(x_{i}).
\end{align}

Then, $b$ is computed using~(\ref{Eq:kkt2}). 
Finally, the decision function with respect to $\textbf{x}$ can be obtained as follows:
\begin{align}
\label{Eq:svm_des}
f(x)=sign(w.\Phi(x)+b)=sign(\sum^{N}_{i=1}\alpha_{i}y_{i}K(x_i,x)+b).
\end{align}

Algorithm~\ref{al:SVM} summarizes the overall procedure of the original SVM model for binary classification problems.

\begin{algorithm} 
\caption{ SVM for Binary Classification.}
\label{al:SVM}
\begin{algorithmic}[1]
\State \textbf{Input:}
\State \hspace{\algorithmicindent} Training dataset: $\{(\textbf{x}_i,y_i)\}_{i=1}^N$ where $\textbf{x}_i \in \mathbb{R}^D$ and $y_i \in \{-1, 1\}$
\State \textbf{Output:}
\State \hspace{\algorithmicindent} SVM model parameters ($w$, $b$)
\State \textbf{Procedure:}
\State Define the feature space transformation $\Phi$.
\State Formulate the optimization problem:
\State \hspace{\algorithmicindent} For linear problems:
\State \hspace{\algorithmicindent}\hspace{\algorithmicindent} $\min(\frac{1}{2}w.w)$ subject to $y_{i}(w.\Phi(x_{i})+b)\geq1$
\State \hspace{\algorithmicindent} For non-linear problems:
\State \hspace{\algorithmicindent}\hspace{\algorithmicindent} $\min(\frac{1}{2}w.w+C\sum_{i=1}^{N}\xi_{i})$ subject to $y_{i}(w.\Phi(x_{i})+b)\geq1-\xi_{i}$ \& $\xi_{i}\geq0$
\State Solve the optimization problem using Lagrangian multipliers:
\State \hspace{\algorithmicindent} $\max\{\sum^{N}_{i=1}\alpha_{i}-\frac{1}{2}\sum^{N}_{i=1}\sum^{N}_{j=1}\alpha_{i}\alpha_{j}y_{i}y_{j}K(x_{i},x_{j})\}$
\State \hspace{\algorithmicindent} subject to $\sum^{N}_{i=1}y_{i}\alpha_{i}=0$ and $0\leq\alpha_{i}\leq C$
\State Compute $w$ using $w=\sum^{N}_{i=1}\alpha_{i}y_{i}\Phi(x_{i})$.
\State Compute $b$ using KKT conditions.
\State Construct the decision function $f(x)$:
\State \hspace{\algorithmicindent} $f(x)=\text{sign}(\sum^{N}_{i=1}\alpha_{i}y_{i}K(x_i,x)+b)$.
\end{algorithmic}
\end{algorithm}

\subsection{Fuzzy support vector machines}
\label{Sec:sec:fsvm}
Let $\{(x_{i},y_{i},s_{i})\}_{i=1}^N$represent a data set with $N$ samples, where $s_i$ ($\sigma\leq s_{i}\leq1$) is the fuzzy membership of $i$-th sample and $\sigma>0$ is a small positive value, while $z=\phi(x)$ indicates a mapping $\phi$ from $R^{D}$ to a feature space $z$~\cite{lin2002fuzzy,Batuwita2010fsvm}. 
The optimal hyperplane is obtained by:
\begin{align}\label{Eq:fsvm1}
min~\frac{1}{2}w^{T}.w+C\sum^{N}_{i=1}s_{i}\xi_{i}
\end{align}
\begin{align*}
s.t.~y_{i}(w.z_{i}+b)\geq1-\xi_{i},~~\xi_{i}\geq0,~i=1,...,N,
\end{align*}

{  The Lagrangian is constructed to solve this problem as follows:}
\begin{align}\label{Eq:fsvm2}
L(w,b,\xi,\alpha,\beta)=\frac{1}{2}w^{T}.w+C\sum^{l}_{i=1}s_{i}\xi_{i}-
\sum^{l}_{i=1}\alpha_{i}(y_{i}(w.z_{i}+b)-1+\xi_{i})-\sum^{l}_{i=1}\beta_{i}\xi_{i}.
\end{align}

To find the saddle point of $L(w,b,\xi,\alpha,\beta)$, the following conditions should be satisfied:
\begin{align}\label{Eq:fsvm_con1}
\frac{\partial L(w,b,\xi,\alpha,\beta)}{\partial w}=w-\sum^{N}_{i=1}\alpha_{i}y_{i}z_{i}=0
\end{align}
\begin{align}\label{Eq:fsvm_con2}
\frac{\partial L(w,b,\xi,\alpha,\beta)}{\partial b}=-\sum^{N}_{i=1}\alpha_{i}y_{i}=0
\end{align}
\begin{align}\label{Eq:fsvm_con3}
\frac{\partial L(w,b,\xi,\alpha,\beta)}{\partial \xi_{i}}=s_{i}C-\alpha_{i}-\beta_{i}=0.
\end{align}

By applying~(\ref{Eq:fsvm_con1}-\ref{Eq:fsvm_con3}) into (\ref{Eq:fsvm1}) and (\ref{Eq:fsvm2}):
\begin{align}\label{Eq:fsvm3}
\textit{maximize}~W(\alpha)=\sum^{N}_{i=1}\alpha_{i}-\frac{1}{2}\sum^{N}_{i=1}\sum^{N}_{j=1}\alpha_{i}\alpha_{j}y_{i}y_{j}K(x_{i},x_{j})
\end{align}
\begin{align*}
s.t.~\sum^{N}_{i=1}y_{i}\alpha_{i}=0,~~0\leq\alpha_{i}\leq s_{i}C,~~i=1,...,N.
\end{align*}
and the KKT conditions can be written as:
\begin{align}\label{Eq:kktfsvm}
\bar{\alpha}_{i}(y_{i}(\bar{w}.z_{i}+\bar{b})-1+\bar{\xi}_{i})=0,~~i=1,...,N
\end{align}
\begin{align*}
(s_{i}C-\bar{\alpha}_{i})\xi_{i}=0,~~i=1,...,N.
\end{align*}
The input sample $x_{i}$ with $\bar{\alpha}_{i}>0$ is denoted as the support vector. 
FSVM contains two types of support vectors: \textit{(i)} support vector with $0<\bar{\alpha}_{i}<s_{i}C$, which lies on the hyperplane margin, and \textit{(ii)} the support vector with $\bar{\alpha}_{i}=s_{i}C$,which indicates misclassification. In general, fuzzy-based SVM models can reduce the sensitivity of conventional SVM toward handling noisy samples.

\subsection{Twin Support Vector Machine}
\label{Sec:sec:tsvm}
TSVM~\cite{khemchandani2007twin} consists of two non-parallel hyperplanes, i.e., a hyperplane with respect to data samples of each class, as follows:
\begin{align}
\label{Eq:tsvm}
w_{1}.x_{i}+b_{1}=0,~~~~w_{2}x_i+b_{2}=0
\end{align}
where $w_{j}$ and $b_{j}$ are the weight vector and bias of the j-th hyperplane, respectively. A quadratic optimization problem is formulated to obtain both hyperplanes, as follows:
\begin{align}\label{Eq:tsvm2}
\underset{w_{1},b_{1},\xi_{2}}{\min}~\frac{1}{2}(Aw_{1}+e_{1}b_{1})^{T}(Aw_{1}+e_{1}b_{1})+p_{1}e^{T}_{2}\xi_{2}
\end{align}
\begin{equation*}
s.t.~~~-(Bw_{1}+e_{2}b_{1})+\xi_{2}\geq e_{2},~\xi_{2}\geq 0,
\end{equation*}
and
\begin{align}\label{Eq:tsvm3}
\underset{w_{2},b_{2},\xi_{1}}{\min}~\frac{1}{2}(Bw_{2}+e_{2}b_{2})^{T}(Bw_{2}+e_{2}b_{2})+p_{2}e^{T}_{1}\xi_{1}
\end{align}
\begin{align*}
s.t.~~~(Aw_{2}+e_{1}b_{2})+\xi_{1}\geq e_{1},~\xi_{1}\geq 0,
\end{align*}
where $A$ and $B$ indicate data samples belonging to classes $+1$ and $-1$, respectively, $e_{1}$ and $e_2$ are vectors of ones, and $p_{1}$ and $p_{2}$ are penalty parameters. 
{  Once the optimal parameters, i.e., $(w^{\ast}_{1},b^{\ast}_{1})$ and $(w^{\ast}_{2},b^{\ast}_{2})$, are achieved, a test sample, $x$, can be classified,} as follows:
\begin{equation}\label{Eq.26}
f(x)=\arg\min_{j~\in\{1,2\}}\frac{\mid (w^{\ast}_{j})^{T}x+b^{\ast}_{j} \mid}{\parallel w^{\ast}_{i} \parallel}.
\end{equation}

\begin{figure}
\centering
    \begin{subfigure}[b]{0.5\textwidth}
            \includegraphics[width=\textwidth]{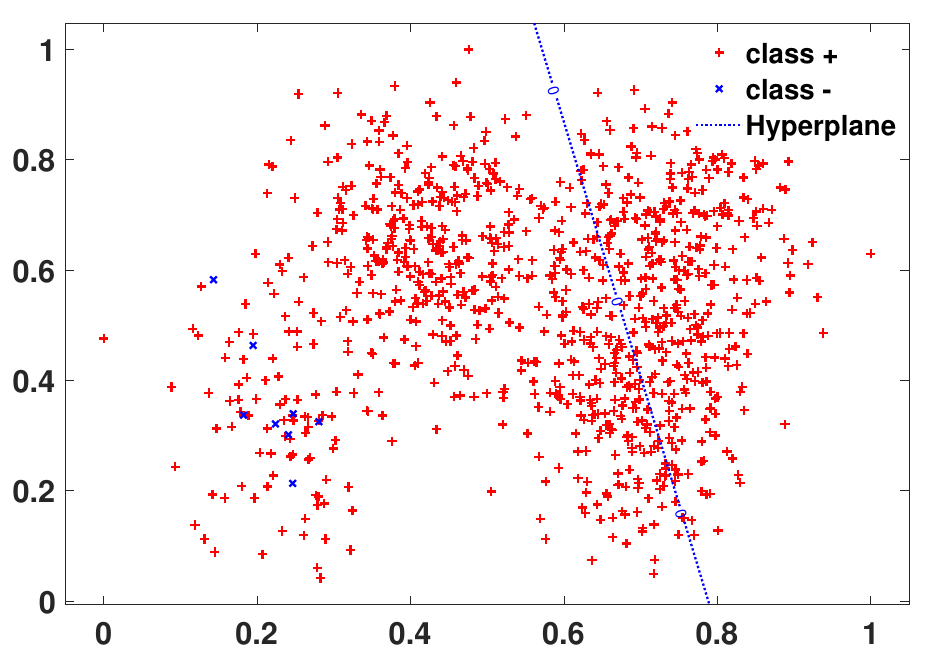}
            \caption{}
            \label{fig:ssvm}
    \end{subfigure}%
    \begin{subfigure}[b]{0.5\textwidth}
            \centering
            \includegraphics[width=\textwidth]{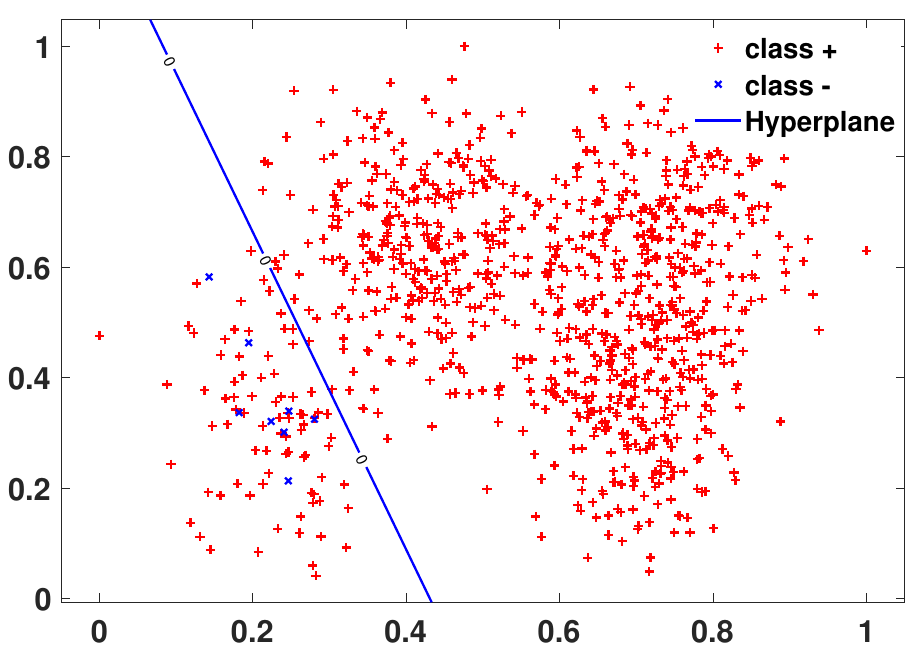}
            \caption{}
            \label{fig:imsvmm}
    \end{subfigure}\\
        \vspace*{0.5cm}
    \caption{{ (a) A hyperplane formed by SVM for an imbalanced data set; (b) an ideal hyperplane that is expected to be formed by SVM-based methods for an imbalanced data set.}}\label{fig:svm}
\end{figure}

\subsection{Reasons for Poor Performance of SVM with Imbalanced Data sets}
\label{secsec:imbsvm}
Fig.~\ref{fig:svm} shows an example of SVM hyperplane (Fig.~\ref{fig:ssvm}) and an ideal hyperplane (Fig.~\ref{fig:imsvmm}) for a class imbalanced data set.
As can be observed in Fig.~\ref{fig:ssvm}, SVM learns a decision boundary close toward the majority class and far away from the minority class~\cite{akbani2004applying,wu2003class}.
The first reason is the distribution of imbalanced data, in which data samples of the minority class are far away from the ideal decision boundary. The second reason is because of the weakness associated with soft margins. As stated earlier, the objective function of SVM aims to \textit{(i)} maximize the margin by minimizing the regularization term, and \textit{(ii)} minimize the total error by minimizing $\sum_{i=1}^N\xi_i$. Constant $C$ specifies the trade-off between maximizing margin and minimizing error. To reduce the penalty, the total error should be reduced, because of assigning the same error cost to all samples by SVM. For an imbalanced data set, a small $C$ makes the margin large, leading to incorrect consideration of the minority class samples as noise. As such, SVM learns all samples as from the majority class with (close to) zero error on negative samples and small error on the minority class samples. This results in learning a decision boundary close (skewed) towards the majority class. One solution to alleviate this issue is to change the trade-off based on the  imbalanced rate~\cite{cristianini2000introduction,veropoulos1999controlling}.\par

The third reason is the imbalanced rate of support vectors~\cite{wu2003class}. When the ratio of imbalanced samples increases, the rate of support vectors with respect to the majority and minority classes becomes more imbalanced. This causes the neighborhood of a test sample near the decision boundary to be dominated by support vectors of the majority class; hence, SVM tends to classify it as a majority class sample. Due to the KKT condition in~(\ref{Eq:kkt1}) and~(\ref{Eq:kkt2}), the summation of $\alpha$'s related to the support vectors of the majority class should be equal to the summation of $\alpha$'s related to those of the minority class. Since the number of support vectors belonging to the minority class is far fewer than those of the majority class, their $\alpha$'s are larger than those of the majority class. These $\alpha$'s operate as weights during prediction. As such, the support vectors of the minority class receive higher weights than those of the majority class, offsetting the effect of support vector imbalanced.\par

\begin{figure}[bt!]
 \begin{center}
 \includegraphics[width=0.9\textwidth]{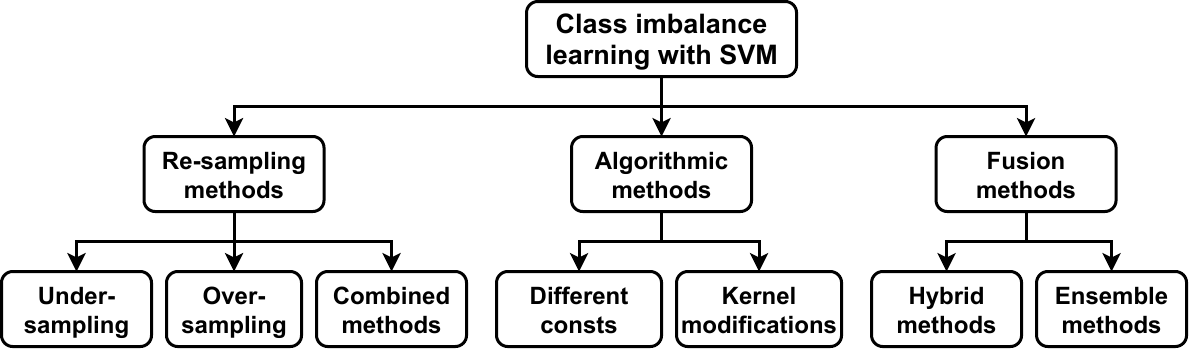}
  \end{center}
  \caption{{ A hierarchical categorization of SVM-based methods for class imbalanced learning.}}
  \label{fig:taxo}
\end{figure}

\section{Imbalanced Support Vector Machine}
\label{Sec:isvm}
\begin{figure}
\centering
    \begin{subfigure}[b]{0.49\textwidth}
            \includegraphics[width=\textwidth]{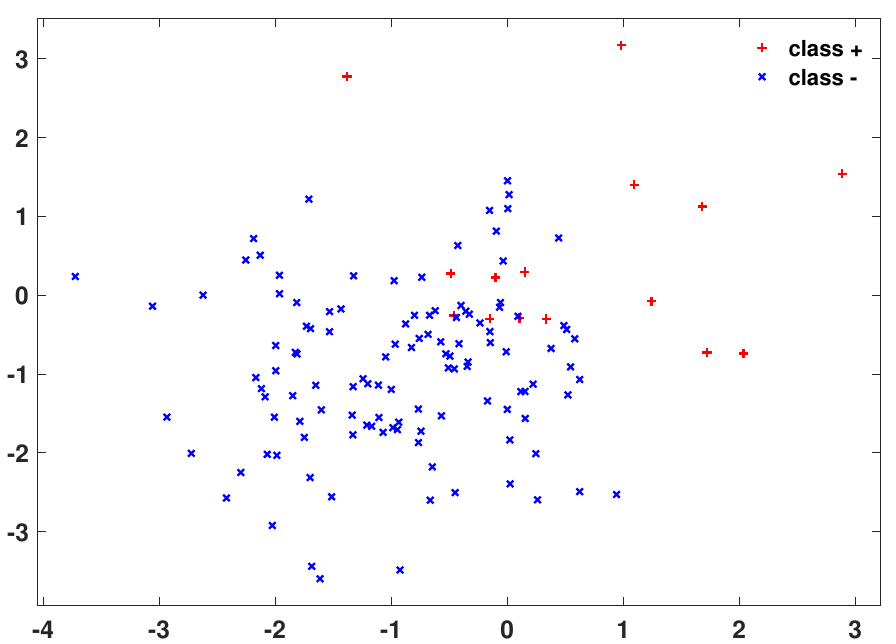}
            \caption{}
            \label{fig:SSL_F1}
    \end{subfigure}%
    \begin{subfigure}[b]{0.49\textwidth}
            \centering
            \includegraphics[width=\textwidth]{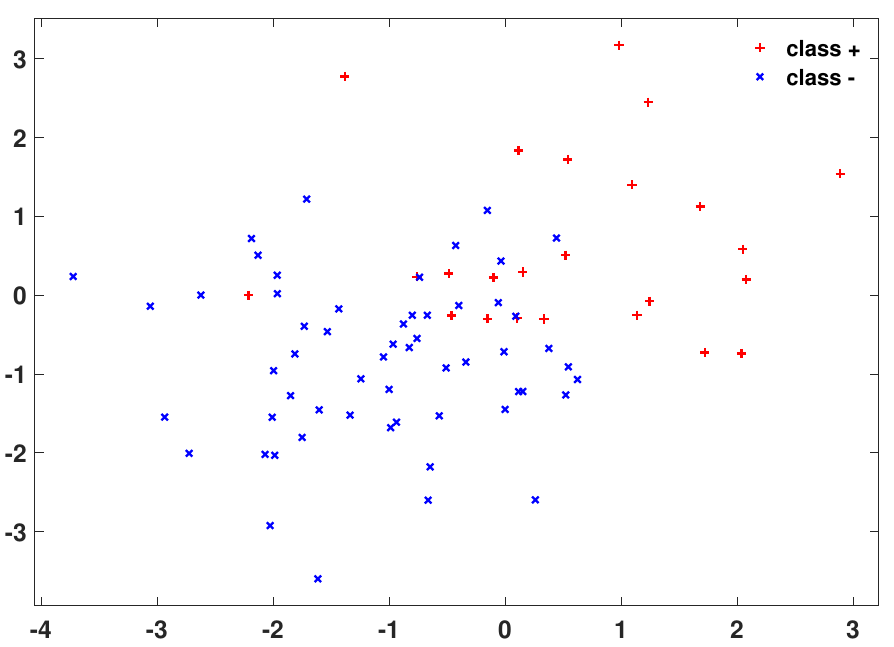}
            \caption{}
            \label{fig:SSL_F2}
    \end{subfigure}\\
    \vspace*{0.5cm}
    \begin{subfigure}[b]{0.49\textwidth}
            \centering
            \includegraphics[width=\textwidth]{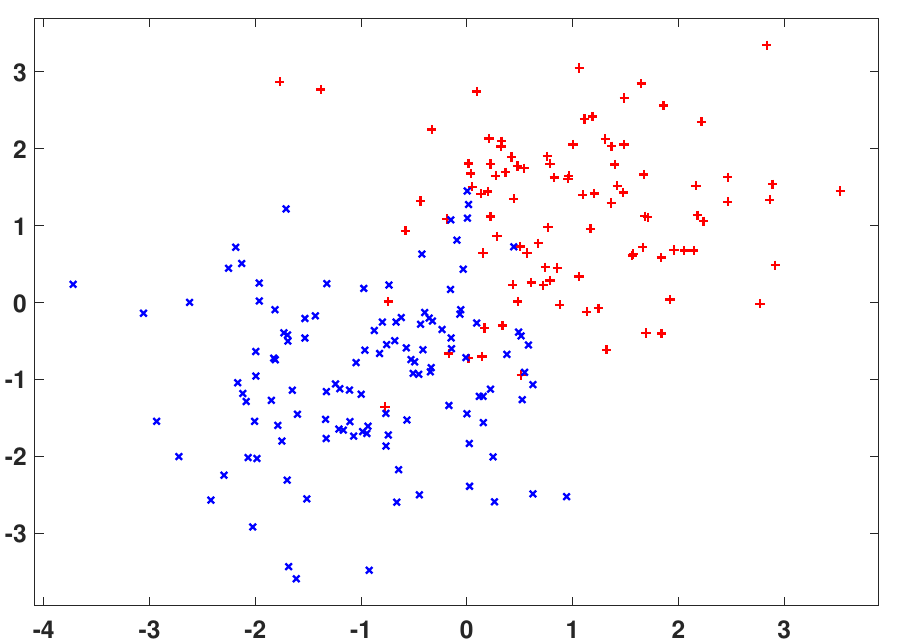}
            \caption{}
            \label{fig:SSL_F3}
    \end{subfigure}
    \begin{subfigure}[b]{0.49\textwidth}
            \centering
            \includegraphics[width=\textwidth]{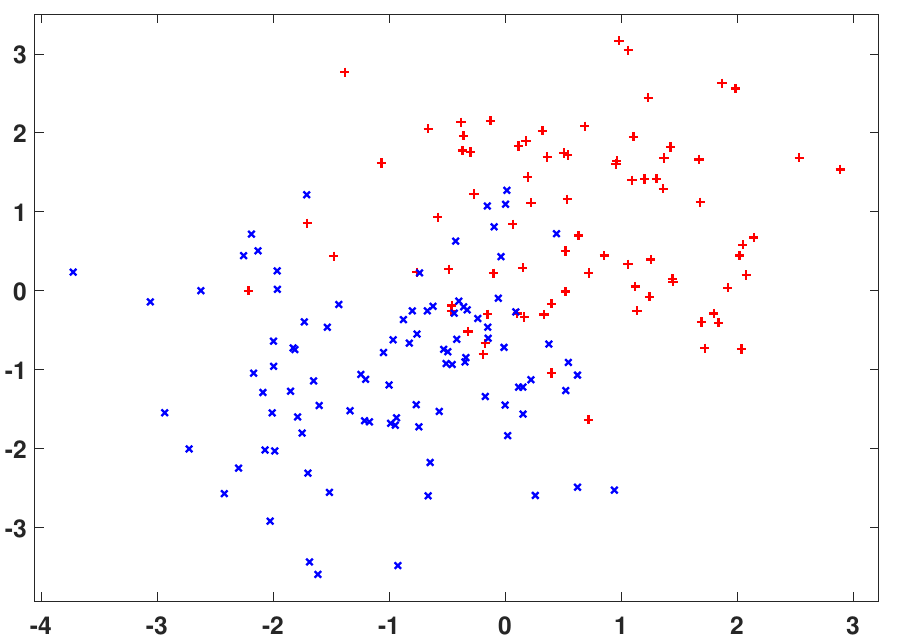}
            \caption{}
            \label{fig:SSL_F4}
    \end{subfigure}
    \vspace*{0.5cm}
    \caption{{ Examples of (a) an imbalanced data set, (b) under-sampling, (c) over sampling, and (d) combined re-sampling.}}\label{fig:ComData_level}
\end{figure}

As stated earlier, the original SVM and its variants are susceptible to imbalanced data issue. Methods to improve SVM capabilities for solving imbalanced data problems have been developed. In general, the existing methods can be grouped into three broad categories (Fig.~\ref{fig:taxo}), as follows:

\begin{itemize}
\item \textbf{\textit{Re-sampling methods}} pre-process the data set and attempt to balance the numbers of samples belonging to different majority and minority classes before training the learning model. This can be achieved by under-sampling the majority class~(Fig.~\ref{fig:SSL_F2}), over-sampling the minority class (Fig.~\ref{fig:SSL_F3}) or a combination of under-sampling and over-sampling (Fig.~\ref{fig:SSL_F4}), as discussed in Sub-section~\ref{Sec:sec:data-level}. 
\item \textbf{\textit{Algorithmic methods}} modify the learning model in a way that penalizes the misclassification rate of the minority class by using different cost-sensitive metrics or modifying the underlying kernel. Sub-section~\ref{sec:sec:alg} presents these methods and introduces the representative models.
\item \textbf{\textit{Fusion methods}} combine various techniques to learn from imbalanced data. The strategies can be fusion of a re-sampling method with an algorithmic method or ensemble method. Sub-section~\ref{sec:sec:hyb} reviews hybrid methods.
\end{itemize}

\vspace{-0.2cm}
\subsection{Re-sampling Methods}
\label{Sec:sec:data-level}
Many studies have shown that most base classifiers, including SVM and its variants, can produce good results in learning with a balanced data set, as compared with an imbalanced one~\cite{estabrooks2004multiple}. The findings indicate the necessity of employing re-sampling methods for learning with imbalanced data. In this regard, many re-sampling methods have been proposed to improve the performance of SVM and its variants in tackling imbalanced data~\cite{chawla2002smote,chen2004classification,fu2004ablock,lessmann2004solving,batuwita2008animproved,batuwita2009micropred,osorio2021relevant}. These methods can be further categorized into two: under-sampling and over-sampling. In the following sub-sections, we discuss both categories and present several representative models of each category. It is noteworthy to mention that after applying re-sampling methods, a variety of algorithms can be used to perform  classification. In this paper, we review studies that employ SVM and its variants as a base classifier. Study~\cite{tyagi2019sampling} evaluated popular re-sampling methods and concluded that under-sampling produces better results.\par

\subsubsection{Under-sampling methods}
\label{Sec:sec:sec:under}
This category covers straightforward methods for handling imbalanced data. The data distribution among different classes is made balanced by selecting fewer samples from the majority class (see Fig.~\ref{fig:SSL_F2}).
According to ~\cite{kubat1997addressing}, the majority class samples can be divided into four groups: \textit{(i)} samples that contain class-label noise, \textit{(ii)} samples that are located along the decision boundary between the majority and minority classes, which are unreliable, \textit{(iii)} redundant samples that can be replaced by other samples, and \textit{(iv)} safe samples which are worth keeping for classification. The study indicated that the classification performance can be improved if safe samples of the majority class along with samples of the minority class are used for training.\par

Over the years, a number of under-sampling techniques have been developed. Among them, random sampling is the most popular one, which randomly selects a subset of majority class samples to train a classifier~\cite{liu2009exploratory,Tang2019Urban,tang2020novel,zheng2021Under}.
In~\cite{liu2008exploratory}, two ensemble strategies were introduced: EasyEnsemble and BalanceCascade. After randomly grouping the majority class samples into several subsets, EasyEnsemble uses each subset to train one classifier, and then combines all classifiers. BalanceCascade uses a trained classifier to guide the sampling process in a way that correctly classified majority class samples are removed. This procedure is repeated until an equal number of samples for both majority and minority classes remain.

Early random sampling methods rely on the single random under-sampling (SRU) strategy. SRU randomly selects majority class samples only once, which could select inappropriate samples, resulting in inferior results. To address this issue, a repeatedly random under-sampling (RRU) strategy was proposed in~\cite{Tang2019Urban}.
RRU sets a threshold, i.e., the area under the receiver operating characteristic curve (AUC) $>$ 0.9,  to enhance the generalization capacity of a classifier during each iteration. However, random sampling methods cannot guarantee good results from the trained classifier. To alleviate this issue, optimized RRU was introduced in~\cite{tang2020novel}. It adopts an AUC comparison among the SRU-based classifiers. 
In~\cite{zheng2021Under}, the majority class samples that could cause misclassification of minority class samples are firstly identified. Then, an under-sampling algorithm, which uses an internal interpolation of several samples to retain sample information, is applied to remove samples from a data set. Besides that, weighted under-sampling (WU)~\cite{kang2018adistance} groups the majority class samples into sub-regions (SRs) and assigns a weight to each SR based on the Euclidean distance with respect to a hyperplane. The samples in SR with a higher weight have a higher chance to be sampled in each iteration.\par

Another popular under-sampling method is data clustering. Data samples are firstly clustered into several groups before utilizing different under-sampling methods~\cite{yuan2006learning,Ng2015Diversified,di2019density,Shukla2017improve,rezvani2021class}. 
As an example, support cluster machines (SCMs)~\cite{yuan2006learning} adopt the $k$-means algorithm to cluster the majority class samples into several disjoint clusters in the feature space. An SVM is then trained using the minority class samples. Using the trained SVM, the support vectors are approximately identified. 
A shrinkage technique is used to remove data samples that are not support vectors. This procedure is repeated until a termination condition is satisfied. In diversified sensitivity-based under-sampling (DSUS)~\cite{Ng2015Diversified}, the data samples from both majority and minority classes are grouped into $k$ clusters separately using the $k$-mean algorithm. In each cluster, the sample closest to the cluster center is identified for both classes, and the similarity measure (SM) score of each sample is computed using the stochastic sensitivity measure. The sample with the highest SM score is selected from each class to form the initial data set for training the base classifier. This procedure is repeated until a balanced set of samples can yield a high sensitivity rate is obtained.\par 

The study~\cite{Wei2017Clustering} introduced two strategies. In the first strategy, the cluster centers are used to represent the majority class samples, while in the second strategy, the nearest neighbours are used to represent them. The study in~\cite{rezvani2021class} adopted the fuzzy adaptive resonance theory (ART) neural network~\cite{carpenter1991fuzzy} for data clustering. Operating as an incremental learning algorithm, Fuzzy ART is used to cluster all training samples and identify clusters that contain only the majority class samples. Then, the cluster centers, instead of data samples, are utilized to train an intuitionistic FTSM (IFTSVM)~\cite{rezvani2019intuitionistic} model for performing classification.\par

In~\cite{ertekin2007active,ertekin2007learning}, an active learning-based selection strategy was developed. Firstly, an SVM is trained using imbalanced data. Then, the closest samples to the hyperplane are identified and added to the training set. This procedure is continued until a termination condition is satisfied.\par

\subsubsection{Over-sampling methods}
\label{Sec:sec:sec:over}
This category of re-sampling methods synthesizes artificial samples for the minority class to balance the numbers of samples between both minority and majority classes (see Fig.~\ref{fig:SSL_F3}).
{ The} synthetic minority over-sampling technique (SMOTE)~\cite{chawla2002smote} generates synthetic samples. It computes the difference between the feature vectors of a sample and its nearest neighbour, and then multiplies the computed difference with a random number and adds it to the feature vector of the sample. SMOTE performs better than random over-sampling techniques, as empirically demonstrated in~\cite{Kerdprasop2011Predicting}. Since then, several enhanced SMOTE algorithms have been developed. The study in~\cite{Jinjin2013new} introduced a graph using the $K$-nearest neighbour for the minority class samples. The most informative samples are identified based on the minimum spanning tree. Then, artificial samples are generated using SMOTE. According to~\cite{Liang2020LR}, LR-SMOTE locates newly generated samples close to the sample centers.  
The study in~\cite{sreejith2020clinical}, an Orchard SMOTE (OSMOTE) model was devised. It uses Orchard’s algorithm to find the nearest neighbours of the minority class samples. Due to its linear nature, SMOTE is ineffective in dealing with non-linear problems. To alleviate this issue, weighted kernel-based SMOTE (WK-SMOTE)~\cite{Mathew2018Kernel} performs over-sampling in the feature space of SVM.\par

Borderline Over-sampling (BOS)~\cite{Nguyen2011Borderline}  performs over-sampling using the minority class samples located near the decision boundary. BOS exploits support the support vectors of an SVM trained with the original training set. It then randomly generates new samples along the trajectory that joins each minority class support vector with a number of its nearest neighbours. The majority weighted minority over-sampling technique (MWMOTE)~\cite{barua2014mwmote} first assigns weights to the hard-to-learn informative minority class samples based on their Euclidean distance from the majority class samples. It then adopts a clustering algorithm to generate the synthetic samples from the weighted minority class samples. Cluster-MWMOTE~\cite{jianan2020complex}, which is an extension of MWMOTE, combines MWMOTE with agglomerative hierarchical clustering to avoid neglecting small subclusters of minority class samples.\par

The study in~\cite{batuwita2010efficient} selected the most informative majority class samples located along the decision boundary. The separating hyperplane is obtained by training an SVM using the original imbalanced data set. Then, a random over-sampling technique is applied to duplicate the number of minority class samples. On the other hand, the synthetic informative minority oversampling (SIMO)~\cite{piri2018asynthetic} algorithm performs over-sampling on the minority class samples located close to the decision boundary. Weighted SIMO (W-SIMO)~\cite{piri2018asynthetic} focuses on the incorrectly classified minority class samples. Accordingly, incorrectly classified minority class samples are over-sampled, as compared with the correctly classified minority class samples. Noise reduction a priori synthetic over-sampling (NRAS)~\cite{William2017Noise} performs noise removal in the minority class samples before generating new samples. In this regard, the $k$ nearest neighbour of each minority class sample is checked. If the neighbour has a value of membership lower than a threshold, the sample is removed. Generative over-sampling~\cite{liu2010Effects} generates data samples using the learned parameters from certain probability distribution. OOLASVM~\cite{Himaja2018Oversample} is a hybrid active sample selection and over-sampling method for learning a new boundary. The sample-characteristic over-sampling technique (SCOTE)~\cite{wei2021new} offers a multi-class imbalanced fault diagnosis method. It transforms a multi-class imbalanced problem into multiple binary problems. Then, SCOTE uses the KNN to remove noisy samples in each binary imbalanced problem. After training using the least squares SVM (LS-SVM), the minority class samples are sorted based on their importance, i.e., misclassification error of the minority class. Finally, the new samples are generated based on the informative samples.\par

\subsubsection{Combined re-sampling methods}
\label{Sec:sec:sec:combined}
{ These methods combine} over-sampling the minority class samples and under-sampling the majority class samples in a pre-processing phase (see Fig.~\ref{fig:SSL_F4}). 
Under-sampling is useful when there are sufficient minority class samples despite an imbalanced ratio. It reduces the burden on storage and requires a shorter execution duration. However, it can lead to the loss of useful information by removing data samples located along the decision boundary. In particular, random under-sampling does not take into consideration informative samples along the decision boundary. Moreover, since the data distribution is normally unknown, most classifiers attempt to approximate it through sample distribution. In under-sampling, the data samples are not considered as random. In this regard, the study in~\cite{akbani2004applying} empirically analyzed the factors associated with this limitation. Explanations on why under-sampling may not be the best choice for SVM are provided.\par

In contrast, over-sampling can result in over-fitting owing to duplicating the available samples or generating new samples based on existing ones. It also increases the computational burden during training. To alleviate these shortcomings, the study in~\cite{chawla2002smote} empirically combined random under-sampling and SMOTE can perform better than under-sampling and over-sampling techniques.  Combining under-sampling and over-sampling reverses the initial bias of the classification algorithm towards the majority class in favour of the minority class.\par

\vspace{-0.2cm}
\subsection{Algorithmic Methods}
\label{sec:sec:alg}
Many modifications from the algorithm perspective have been developed to enable SVM and its variants for learning imbalanced data~\cite{ganaie2020joint,yanze2020quasi,cao2020expediting,liu2020fuzzy,tanveer2020least,wu2020joint,cao2013anoptimized}. The key modifications are discussed in the following sub-sections.  

\subsubsection{Different error costs (DEC)}
\label{sec:sec:sec:dec}
This category of algorithmic methods uses different cost functions for the majority and minority class samples, as follows~\cite{veropoulos1999controlling,bach2006considering}: 
\begin{eqnarray}
min\frac{1}{2}w.w+(C^{+}\sum^{l}_{i\mid y_{i}=+1}\xi_{i}+C^{-}\sum^{l}_{i\mid y_{i}=-1}\xi_{i})
\end{eqnarray}
\begin{equation}\label{Eq.33}
s.t.~y_{i}(w.\Phi(x_{i})+b)\geq1-\xi_{i}~\xi_{i}\geq0~i=1,2,...,l
\end{equation}
where $C^{+}$ and $C^{-}$ indicate the error rates of the minority and majority classes, respectively. The formulation in~(\ref{Eq.33}) reduces the impact of class imbalanced by assigning a higher error rate for the minority class samples than that of majority class samples~\cite{veropoulos1999controlling}. 
 It prevents skewing the separating hyperplane to the minority class samples. The associated Lagrangian function is as follows: 
\begin{equation}\label{Eq.599}
max\{\sum^{l}_{i=1}\alpha_{i}-\frac{1}{2}\sum^{l}_{i=1}\sum^{l}_{j=1}\alpha_{i}\alpha_{j}
y_{i}y_{j}K(x_{i},x_{j})\}
\end{equation}
\begin{eqnarray*}\label{Eq.34}
s.t.~\sum^{l}_{i=1}y_{i}\alpha_{i}=0,~0\leq\alpha_{i}^{+}\leq C^{+},~0\leq\alpha_{i}^{-}\leq C^{-}~i=1,...,l
\end{eqnarray*}
where $\alpha_{i}^{+}$ ($\alpha_{i}^{-}$) describes the Lagrangian multipliers of minority (majority) class samples. A good classification performance can be achieved by tuning $C^{-}/C^{+}$. However, there is no defined indication pertaining to the values that should be selected for the penalty term.  As an example, in Kowalczk and Raskutti~\cite{kowalczyk2002one,raskutti2004extreme} set $C^{+}$ and $C^{-}$ are set to $C(1-B)/n_{+}$ and $CB/n_{-}$, respectively. Where $n_{+}(n_{-}$) indicates the number of minority (majority) class samples, $C>0$ is a regularization constant, and $0\leq B\leq 1$ is the balance factor. Note that $B=0$ indicates the extreme case of learning from the minority class samples only (one-class learning), and vice versa.      

zSVM~\cite{iman2006zsvm} focuses on adjusting the learned hyperplane to obtain an optimal margin between the decision boundary separating the majority and minority classes. To achieve this, the general decision function, i.e.,~(\ref{Eq:svm_des}), is changed to:  
\begin{multline}\label{Eq.36}
f(x)=sign(\sum^{l}_{i=1}\alpha_{i}y_{i}K(x,x_{i})+b)\\
=sign(\sum^{n_{+}}_{i=1}\alpha_{i}^{+}y_{i}K(x,x_{i})+\sum^{n_{-}}_{j=1}\alpha_{j}^{-}y_{j}K(x,x_{j})+b),
\end{multline}
where $\alpha_{j}^{+}$ ($\alpha_{i}^{-}$) is the coefficient of the positive (negative) support vectors. zSVM increases $\alpha_{i}^{+}$ in the positive support vectors by multiplying them with a positive value, $z$. The resulting decision function in~(\ref{Eq.5779}) reduces bias towards the majority class samples. The weights of the positive support vectors in ~(\ref{Eq.5779}) are increased by adjusting $\alpha_{i}^{+}$.
\begin{equation}\label{Eq.5779}
f(x)=sign(z\ast\sum^{l_{1}}_{i=1}\alpha_{i}^{+}y_{i}K(x,x_{i})+\sum^{l_{2}}_{j=1}\alpha_{i}^{-}y_{i}K(x,x_{j})+b).
\end{equation}

Several studies devise cost-sensitive SVM (CS-SVM) models for learning from imbalanced data sets. The study in~\cite{gu2017cross} formulated the error surfaces based on cross-validation (CV-SES). Specifically, a bi-parameter space partition algorithm computes a two-dimensional solution surface for CS-SVM, which can fit the CS-SVM solutions for the regularization parameters. Then, a two-dimensional validation error surface for each CV fold is computed, which can fit the CS-SVM validation errors for the regularization parameters. Then, the global minimum CV error of CS-SVM is found by obtaining the CV error surface through superposing $K$ validation error surfaces.
The study in~\cite{yang2009margin} used an inverted proportional regularized penalty to change the weight of imbalanced data. CSRankSVM~\cite{yu2020improving}, which is a cost-sensitive ranking SVM, improves the performance of ranking-oriented defect predictions. The studies in~\cite{xu2018cnc,ma2020sincremental} proposed incremental learning models using cost-sensitive SVM models to tackle imbalanced classification problems in online learning environments.  
In addition, the study in~\cite{Sebastian2018Embedded} developed a feature selection strategy to penalize the cardinality of the feature set using a scaling factor technique. Two variants of SVM, i.e, CS-SVM and support vector data description~\cite{tax2004support}, are utilized to handle high-dimensional class imbalanced data.

Due to less sensitivity to support vectors and noisy samples, fuzzy-based SVM models are leveraged to tackle imbalanced classification problems. A membership function is assigned to each class according to the importance of different training samples~\cite{Batuwita2010fsvm,lin2002fuzzy}, i.e.,  
\begin{equation}\label{Eq.222}
s_{i}^{+}=f(x_{i}^{+})r^{+},
\end{equation}
\begin{equation}\label{Eq.3}
s_{i}^{-}=f(x_{i}^{-})r^{-},
\end{equation}
where $s_{i}^{+}$ ($s_{i}^{-}$) indicates the membership values of the minority (majority) class samples, $0\leq f(x_{i})\leq 1$ indicates the significance of $x_{i}$ in its class, while $r^{-}$ and $r^{+}$ indicate class imbalanced setting. The misclassification cost for minority (majority) class samples is set to $s_{i}^{+}C$ ($s_{i}^{-}C$), which allocates a value in $[0,1]$ ($[0,r]$), and $r<1$. 

 Over the years, various functions have been introduced, e.g.~(\ref{Eq.222}) and~(\ref{Eq.3}), based on the distance of each sample from the class center~\cite{Batuwita2010fsvm} and hyperplane~\cite{lin2004wang}.
In addition, the study in~\cite{Batuwita2010fsvm} defined $f(x_{i})$ according to the distance from the class center. Eqs.~(\ref{Eq.4}) and (\ref{Eq.37}) present $f(x_{i})$ for linear and non-linear cases, respectively.  
\begin{equation}\label{Eq.4}
f_{lin}(x_{i})=1-(\frac{d_{i}^{cen}}{max(d_{i}^{cen})+\delta}),
\end{equation}
where $\delta$ is a small positive value, ${d_{i}^{cen}}$ is the Euclidean distance of $x_i$ from its corresponding class center. 
\begin{equation}\label{Eq.37}
f_{exp}(x_{i})=\frac{2}{1+exp(d_{i}^{cen}\ast\beta)},
\end{equation}
where $\beta\in[0,1]$ defines the decay steepness.\par

{ The study}~\cite{ma2011anew} improved a bilateral-weighted fuzzy support vector machine (B-FSVM)~\cite{wang2005anew} to tackle imbalanced classification problems. Denoted as B-FSVM-CIL, different membership functions are defined for the minority and majority classes to overcome class imbalanced issues. An entropy-based fuzzy SVM~\cite{fan2017entropy} leverages the entropy measure of class certainty by computing the probability of a sample belonging to each class. The study in~\cite{Gupta2019fuzzy} proposed an entropy-based FTSVM, known as EFTSVM-CIL, for solving binary class imbalanced problems.\par

\subsubsection{Kernel modifications}
\label{sec:sec:sec:kernel}
This category of algorithmic methods makes the SVM models less sensitive to imbalanced data by modifying the kernel function. The available methods can be categorized into three groups~\cite{wu2003adaptive}.
The first group works by aligning the decision boundary~\cite{wu2003class,wu2005kba}. 
Adaptive conformal transformation (ACT)~\cite{wu2003adaptive,wu2003class} changes the spatial resolution around the decision boundary using conformal transformation (CT)~\cite{amari1999improving}. CT maps elements $X\in D$ to elements $Y\in T(D)$ in a way that preserves the local angles between elements after mapping. It adaptively controls the transformation based on the skewness of the decision boundary, i.e., the kernel matrix is modified when the data samples do not have a vector-space representation.\par

The second group leverages on kernel alignment~\cite{cristianini2002on,kandola2003refining}. The class probability-based FSVM (ACFSVM)~\cite{xinmin2020affinity} model uses affinity computed by the support vector description domain (SVDD) to identify outliers and data samples in the majority class that are close to the decision boundary. The kernel k-nearest neighbour algorithm computes the class probability of all majority class samples. ACFSVM reduces the contribution of data samples with lower class probabilities, i.e., noisy samples, and focuses on the majority class samples with higher affinities and class probabilities, therefore skewing the decision boundary toward the majority class. Enhanced automatic twin support vector machine (EATWSVM)~\cite{jimenez2020enhanced} learns a kernel function through Gaussian similarity and improves data separability by incorporating a centered alignment-based technique. The study in~\cite{kandola2003refining} extended the kernel alignment technique~\cite{cristianini2002on} to measure the degree of agreement between a kernel and a classification task. The definition of alignment is modified using different weights for the majority and minority class samples. Other kernel alignment methods include regularized orthogonal weighted least squares (ROWLSs))~\cite{hong2007kernel}, kernel neural gas (KNG)~\cite{qin2004kernel} and P2P communication paradigm (P2PKNNC)~\cite{yu2007novel} to tackle data imbalanced issues.\par

{ Third group} focuses on margin alignment. Maximum margin TSVM (MMTSSVM) \cite{xu2017maximum} identifies two homocentric spheres for tackling imbalanced data. It uses a small sphere to capture as many majority class samples as possible, and increases the distance between two margins to separate most minority class samples using a large sphere.\par

\vspace{-0.2cm}
\subsection{Fusion Methods}
\label{sec:sec:hyb}
Many fusion methods for handling class-imbalanced data have been proposed. They can be broadly divided into two: hybrid and ensemble methods.\par

\subsubsection{Hybrid techniques}
\label{sec:sec:sec:hyb}
This category combines SVM and its variants with other techniques for learning from class imbalanced data. As an example, SMOTE with Different Costs (SDC)~\cite{akbani2004applying} combines SMOTE and the method in~\cite{veropoulos1999controlling}, and changes the Lagrangian function in~(\ref{Eq:tsvm2}) to:
\begin{multline}\label{Eq.559}
L_{p}=\frac{\|w\|^{2}}{2}+C^{+}\sum^{n_{+}}_{\{i|y_{i}=+1\}}\xi_{i}+C^{-}\sum^{n_{-}}_{\{j|y_{j}=-1\}}\xi_{j}\\
-\sum^{n}_{i=1}\alpha_{i}[y_{i}(w.x_{i}+b)-1+\xi_{i}]-\sum^{n}_{i=1}r_{i}\xi_{i},
\end{multline}
With the following conditions:

(1)~If $y_{i}=+1$, then $0\leq\alpha_{i}\leq C^{+}$,

(2)~If $y_{i}=-1$, then $0\leq\alpha_{i}\leq C^{-}$.\\
where $\xi_{i}\geq0$~\cite{cristianini2000introduction}. Thus, the non-zero error cost on negative/positive support vectors  leads to a smaller/larger $\alpha_{i}$. This results in pushing the class boundary toward the direction of the majority class samples. To better establish the class boundary, SMOTE is used to make the minority class samples more densely distributed.\par

A hybrid model for feature selection with maximal between-class separability and a variant of SVM (CCSVM), i.e. feature vector regression~\cite{liu2016feature}, was proposed in~\cite{liu2019applied}. 
In addition, NN-CSSVM was devised in~\cite{hye2020hybrid}. It is a hybrid model of neural networks and CSSVM for handling class imbalanced data with multiple modalities. In~\cite{rtayli2020enhanced}, a hybrid model of recursive feature elimination (RFE) and SMOTE was proposed for credit card fraud detection. This model uses the RFE technique to select an optimal subset of features and SMOTE to overcome the class imbalanced problem. In~\cite{wu2021detection}, a two-stage hybrid screening rule based on variational inequality and duality gap was introduced to accelerate the learning speed of nonparallel SVM~\cite{tian2014nonparallel}.

\subsubsection{Ensemble methods}
\label{Sec:sec:sec:ens}
Ensemble methods train multiple ($M$) learning models $f_i: X\to Y$ for $i=1, \ldots, M$ to perform prediction. The outputs are combined using different techniques to deduce the final decision. In regression, averaging is the simplest ensemble technique for combining the predictions of  $M$ models $f_{ave}: X\to Y$, as follows: 
\begin{equation}\label{simple averaging}
    f_{ave}(x)= \frac{1}{M}\sum\limits_{i=1}^{M}f_i(x).
\end{equation}

Majority voting is a common scheme for fusing the predictions of multiple classifiers to reach an output class,$c=1, \ldots, K$, as follows:
\begin{align}
    f_{majority}(x)= \mathop{\arg\max}\limits_{c} \sum\limits_{i=1}^{M} I(f_i(x)=c).
\end{align}

Bagging~\cite{breiman1996bagging} and boosting~\cite{freund1995boosting} are two popular ensemble methods. Bagging trains $M$ independent classifiers on $M$ different subsets of randomly drawn samples from the original data set.
In~\cite{yua2018adbn}, bagging was used to generate several re-sampled ensemble input data and training subsets. The training subsets are used to train a number of base classifiers, i.e., SVM, while ensemble input data samples are employed by all base classifiers. Then, a deep belief network (DBN)~\cite{hinton2006afast} is adopted to produce the final output by combining the decisions generated by an ensemble of base classifiers.\par

Boosting aims to train a classifier and correct its errors in previous iterations. Adaptive boosting (AdaBoost)~\cite{freund1996experiments} and its variants~\cite{schapire1999improved,sun2020class,mehmood2020customizing} have been extensively used with SVM to deal with class imbalanced problems. The data samples are re-weighted at each training iteration of a classifier. Specifically, it reduces the weights of correctly classified samples, and vice versa. Therefore, the classifier pays more attention to the misclassified samples from the previous iteration. Other useful boosting techniques for tackling class imbalanced problems include Adacost~\cite{fan1999adacost}, RareBoost~\cite{joshi2001evaluating}, and SMOTEBoost~\cite{chawla2003smoteboost}.\par

Over the years, many ensemble models that adopt SVM or its variants as the base classifiers have been devised to deal with class imbalanced problems. Hybrid kernel machines into an ensemble (HKME)~\cite{li2006hybrid} consists of two base classifiers, namely SVM and \textit{v}SVC~\cite{bernhard2001estimating}, which is a one-class SVM classifier. HKME trains SVM with a balanced data set achieved by re-sampling and trains \textit{v}SVC with the majority class samples. Then, an averaging strategy is used to combine the decisions for reaching the final decision. the study in~\cite{bartosz2015hypertension} proposed an ensemble one-class classifier~\cite{tax2004support} for hypertension detection, i.e., a multi-class imbalanced problem. A one-class classifier is trained to identify the majority class samples from outliers. The original multi-class problem is decomposed using an ensemble of one-class classifier to distinguish between five types of hypertension. Then, the error-correcting output codes~\cite{wilk2012soft} is employed to reconstruct the multi-class problem from multiple single-class responses.\par

\begin{table}[tb!]   
\centering
\caption{\label{Table:data} Details of the UCI data sets, including the numbers of samples for the minority and majority classes, total numbers of samples, dimensions, and imbalanced ratio.} 
\begin{adjustbox}{width=0.99\textwidth}
    \begin{tabular}{l l c c c c c }
    \toprule
Data \# & Data set&Minority&Majority&Instance&Dimension&Imbalance ratio\\
\midrule
1& Ripley&650&650&1,250&3&1\\
2& Wisconsin&239&444&683&10&1.86\\
3& Pima&268&500&768&9&1.87\\
4& Yeast 1&429&1,055&1,484&9&2.64\\
5& Vehicle 2&218&628&864&19&2.88\\
6& Vehicle 1&217&629&864&19&2.90\\
7& Transfusion&178&570&748&5&3.20\\
8& Wpbc&47&151&198&34&3.22\\
9& Segment&329&1979&2,308&19&6.02\\
10& Yeast 3&163&1,321&1,484&9&8.10\\
11& Page blocks&560&4,913&5,473&11&8.78\\
12& Yeast 2-vs-4&463&51&514&9&9.08\\
13& Ecoli 0-2-3-4-vs-5&20&182&202&8&9.10\\
14& Yeast 0-3-5-9-vs-7-8&50&456&506&9&9.12\\
15& Yeast 0-2-5-6-vs-3-7-8-9&99&905&1,004&9&9.14\\
16& Ecoli 0-4-6-vs-5&20&183&203&8&9.15\\
17& Ecoli 0-1-vs-2-3-5&24&220&244&8&9.17\\
18& Yeast 0-5-6-7-9-vs-4&51&477&528&9&9.35\\
19& Vowel&90&898&988&11&9.98\\
20 & Ecoli 0-6-7-vs-5&20&200&220&7&10\\
21& Led7digit 0-2-4-5-6-7-8-9-vs-1&37&406&443&8&10.97\\
22& Ecoli 0-1-vs-5&20&220&240&7&11\\
23& Ecoli 0-1-4-7-vs-5-6&25&307&332&7&12.28\\
24& Ecoli 0-1-4-6-vs-5&20&260&280&7&13\\
25& Shuttle c0-vs-c4&123&1,706&1,829&10&13.87\\
26& Glass 4&201&13&214&10&15.46\\
27& Ecoli 4&20&316&336&8&15.80\\
28& Yeast 1-4-5-8-Vs-7&30&663&693&8&22.10\\
29& Glass 5&9&205&214&10&22.78\\
30& Yeast 2-Vs-8&20&462&482&9&23.10\\
31& Yeast 4&51&1,433&1,484&9&28.10\\
32& Yeast 1-2-8-9-Vs-7&30&917&947&9&30.57\\
33& Yeast 5&44&1,440&1,484&8&32.73\\
34& Ecoli 0-1-3-7-vs-2-6&7&274&281&8&39.14\\
35& Yeast 6&35&1,449&1,484&9&41.40\\
36& Abalone 19&32&4,142&4,174&8&129.44\\
\bottomrule
     \end{tabular}
\end{adjustbox}
\end{table}

{ The study in~\cite{zheng2021cardiotocography} developed} an ensemble model based on CS-SVM for recognizing cardiotocography signal abnormality. Specifically, weighting, selecting, and voting techniques are used for producing predictions by combining the outcomes of ensemble members. EasyEnsemble~\cite{liu2006exploratory,liu2009exploratory} and methods in~\cite{lin2009several,kang2006eussvm,liu2006boosting,wang2010boosting,liu2020ensembling,lei2016boosted,li2017instance,tashk2007boosted,Badrinath2016Estimation} divide the majority class samples into several sub-samples that contain the same number of samples as that of the minority class samples using different strategies, e.g. random sampling, clustering or bootstrapping. Then, several SVM classifiers are trained and ensemble techniques are used to  yield the prediction.\par

 The study in~\cite{Maciej2015Boosted} leveraged an active learning strategy for training boosting SVM to exploit the most informative samples. Each ensemble member is trained using data samples that have significant impact on the previous classifier.  Specifically, data samples located along the decision boundary are considered. The penalty terms are computed based on local cardinalities of the majority and minority class samples.\par

\section{Empirical Studies}
\label{Sec:emp}
In this section, two experiments are conducted to compare the performance of various SVM-based models for imbalanced data learning in binary classification settings. The first evaluates different algorithmic methods, while the second assesses several re-sampling and fusion methods. A total of 36 benchmark classification problems from the UCI\footnote{http://
archive.ics.uci.edu/ml/datasets.html} and KEEL\footnote{https://sci2s.ugr.es/keel/imbalanced.php} repositories are used. As shown in Table~\ref{Table:data}, these data sets contain imbalanced ratios from 1 to close to 130. Multi-class data sets are converted into binary classification tasks, as shown in Table~\ref{Table:data}. All data samples are normalized between 0 and 1, and the results of five-fold cross-validation are reported.\par

The SVM parameters are set as follows. The best $C$ value is determined using a grid search scheme covering $\{2^{i}|i=-10,-9,...,0,...,9,10 \}$, $\epsilon$ which is a small value, is selected from ${0, 0.1, 0.2, 0.3}$. The optimal Gaussian kernel function is utilized to deal with non-linear cases, i.e., $\mathcal{K}(x_{1},x_{2})=exp(-\parallel x_{1}-x_{2}\parallel^{2}/\sigma^{2})$ and $\sigma\in\{2^{\sigma_{\min}:\sigma_{\max}}\}$ with $\sigma_{\min}=-10$, $\sigma_{\max}=10$.

The geometric mean (G-mean) and AUC are computed for performance evaluation. G-means measures the trade-off between the classification performance of both majority and minority classes, while AUC indicates the ability of a classifier in distinguishing between the majority and minority classes.\par

\begin{table*}[tb!]   
\centering
\caption{\label{Table:auclin} AUC (\%) rates of SVM-based models for imbalanced data sets with a linear kernel function.}
\begin{adjustbox}{width=0.99\textwidth}
    \begin{tabular}{l c c c c c c }
    \toprule
Data \# &TSVM\cite{khemchandani2007twin}&FTSVM\cite{gao2015coordinate}&EFSVM\cite{fan2017entropy}&FSVM-CIL\cite{Batuwita2010fsvm}&zSVM\cite{Imam2006zSVM}&EFTWSVM-CIL\cite{Gupta2019fuzzy}\\
\midrule
1& 88.12&88.31&83.29&\textbf{89.59} &85.53 &88.53\\
2& 95.32&96.23&\textbf{97.11}&96.87 & 93.56&95.81\\
3& 73.18&73.01&66.89&74.99 &72.51 &\textbf{75.42}\\
5&96.12&93.88&92.48&\textbf{97.30} &95.69 &96.37\\
6&78.31&80.40&65.46&\textbf{82.13} & 79.08&79.95\\
7&49.01&49.01&49.01& \textbf{54.82}&52.32 &52.34\\
8&65.12&64.41&\textbf{69.84}&68.14 &62.67 &67.14\\
12&87.91&89.88&81.53&\textbf{90.04} &85.98 &87.91\\
13&95.24&\textbf{96.76}&92.15&96.57 &94.28 &\textbf{96.76}\\
17&88.36&88.36&70.01& \textbf{88.82}&84.80 &88.36\\
18&76.15&78.80&71.89&78.97 &75.34 &\textbf{79.64}\\
19&91.67&90.62&83.44& \textbf{93.76}&89.57 &91.67\\
21&87.91&86.56&\textbf{90.47}&88.20 & 84.37&88.12\\
22&94.31&94.31&86.84 &92.22 &88.28 &\textbf{94.35}\\
24&\textbf{97.86}&\textbf{97.86}&87.61&97.40 &92.81 &\textbf{97.86}\\
25&\textbf{100}&\textbf{100}&99.32& 99.98& 97.47&\textbf{100}\\
33&95.48&\textbf{96.76}&70.18&95.45 & 91.11&\textbf{96.76}\\
34&94.54&91.31&85.77& \textbf{95.36}&90.41 &94.54\\
\midrule
Average &86.37 & 86.47&80.18 &\textbf{87.81} &84.21 &87.30\\
\bottomrule
     \end{tabular}
\end{adjustbox}
\end{table*}

\subsection{Performance comparison among algorithmic methods}
\label{Sec:sec:exp}
In this experiment, six SVM-based models, namely TSVM~\cite{khemchandani2007twin}, FTSVM~\cite{gao2015coordinate}, EFSVM~\cite{fan2017entropy}, FSVM-CIL~\cite{Batuwita2010fsvm}, zSVM~\cite{Imam2006zSVM} and EFTWSVM-CIL~\cite{Gupta2019fuzzy}, are employed for evaluation. TSVM~\cite{khemchandani2007twin} and FTSVM~\cite{gao2015coordinate} are originally proposed for tackling balanced data while EFSVM~\cite{fan2017entropy}, FSVM-CIL~\cite{Batuwita2010fsvm}, zSVM~\cite{Imam2006zSVM} and EFTWSVM-CIL~\cite{Gupta2019fuzzy} are algorithmic methods for handling imbalanced data. To have a fair comparison, the same parameters are set for all models and two kernel functions, i.e., linear and nonlinear, are used.

\begin{table*}[tb!]   
\centering
\caption{\label{Table:gmanlin} G-Mean (\%) rates of SVM-based models for imbalanced data sets with a linear kernel function.} 
\begin{adjustbox}{width=0.99\textwidth}
    \begin{tabular}{l c c c c c c }
    \toprule
Data \#&TSVM\cite{khemchandani2007twin}&FTSVM\cite{gao2015coordinate}&EFSVM\cite{fan2017entropy}&FSVM-CIL\cite{Batuwita2010fsvm}&zSVM\cite{Imam2006zSVM}&EFTWSVM-CIL\cite{Gupta2019fuzzy}\\
\midrule
1&85.68&85.87&80.99& \textbf{87.11}& 83.16& 86.08\\
2&93.31&94.20&95.06&\textbf{95.83} &91.49 & 93.79\\
3&73.02&72.85&66.74&\textbf{75.68} & 72.25& 75.25\\
5&89.69&87.60&86.29&\textbf{92.54} & 88.35& 89.92\\
6&65.99&67.75&55.16&\textbf{72.58} &69.14 & 67.37\\
7&56.90&56.95&56.95& \textbf{67.33}& 64.28& 60.82\\
8&62.45&61.77&\textbf{66.98}& 66.28&63.26 & 64.39\\
12&84.59&86.49&78.45&\textbf{86.64} &82.73 & 84.59\\
13&94.68&96.19&91.61& 96.23&91.89 & \textbf{96.32}\\
17&91.98&91.95&72.85&\textbf{92.42} & 88.25& 91.95\\
18&75.84&78.45&71.66&78.62 &75.01 & \textbf{79.29}\\
19&87.41&86.41&79.56&\textbf{89.43} &85.39 & 87.41\\
21&89.29&87.92&\textbf{91.89}&89.58&85.50 & 89.50\\
22&\textbf{94.31}&\textbf{94.31}&86.84&92.18 &88.13 &\textbf{94.31}\\
24&92.80&\textbf{92.96}&83.22& 92.41& 88.09& 92.85\\
25&99.96&\textbf{99.98}&99.32&99.97 & 95.44& \textbf{99.98}\\
33&73.09&73.94&86.37&95.63 & 91.27& \textbf{96.94}\\
34&98.09&98.16&89.05&\textbf{98.17} &93.73 & 98.01\\
\midrule
Average &83.84 & 84.10& 79.94&\textbf{87.15} & 83.19& 86.04\\
\bottomrule
     \end{tabular}
\end{adjustbox}
\end{table*}

\begin{table*}[tb!]   
\centering
\caption{\label{Table:aucnonlin} AUC (\%) rates of SVM-based models for imbalanced data sets with a nonlinear kernel function.} 
\begin{adjustbox}{width=0.99\textwidth}
    \begin{tabular}{l c c c c c c }
    \toprule
Data set&TSVM\cite{khemchandani2007twin}&FTSVM\cite{gao2015coordinate}&EFSVM\cite{fan2017entropy}&FSVM-CIL\cite{Batuwita2010fsvm}&zSVM\cite{Imam2006zSVM}&EFTWSVM-CIL\cite{Gupta2019fuzzy}\\
\midrule
1&88.69&88.71&84.37& \textbf{89.62}& 86.54& 88.97\\
2&96.01&96.59& \textbf{97.84}&97.12 & 94.07& 96.54\\
3&73.87&73.43&67.35&75.28 & 72.78&\textbf{75.69}\\
5&96.53&94.27&92.84&\textbf{97.63} &96.04 &96.89\\
6&78.74&81.15&65.93&\textbf{82.78} &79.58 &80.45\\
7&50.52&50.47&50.39&\textbf{55.16} &53.38 &53.68\\
8&65.80&65.10&\textbf{70.43}& 68.97&63.42 &68.01\\
12&88.38&\textbf{90.46}&82.09&90.87 & 86.67&88.56\\
13&95.87&97.14&93.58& 97.19& 95.07&\textbf{97.41}\\
17&88.89&88.87&71.69& 88.98& 85.73&\textbf{89.02}\\
18&76.77&79.55&72.36& 79.47& 76.30&\textbf{80.23}\\
19&92.16&91.72&84.18&\textbf{94.29} &90.35&92.49\\
21&88.36&86.85&\textbf{91.24}&89.39 &85.44&89.40\\
22&\textbf{94.75}&94.42&87.28&92.89 &89.01 &94.61\\
24&98.14&98.27&88.36& 97.82&93.79 &\textbf{98.47}\\
25&\textbf{100}&\textbf{100}&99.76& 99.99& 98.18&\textbf{100}\\
33&96.23&\textbf{97.42}&82.13& 95.88&92.41 &97.36\\
34&95.11&91.85&86.34& \textbf{95.67}&91.29 &95.30\\
\midrule
Average &86.93 &87.01 &81.56 &\textbf{88.26} & 85.01&87.95 \\
\bottomrule
     \end{tabular}
\end{adjustbox}
\end{table*}


\begin{table*}[tb!]   
\centering
\caption{\label{Table:gmeannon} G-Mean (\%) rates of SVM-based models for imbalanced data sets with a nonlinear kernel function. }
\begin{adjustbox}{width=0.99\textwidth}
    \begin{tabular}{l c c c c c c }
    \toprule
Data set&TSVM\cite{khemchandani2007twin}&FTSVM\cite{gao2015coordinate}&EFSVM\cite{fan2017entropy}&FSVM-CIL\cite{Batuwita2010fsvm}&zSVM\cite{Imam2006zSVM} & EFTWSVM-CIL\cite{Gupta2019fuzzy}\\
\midrule
1&86.49&86.08&82.16& \textbf{87.97}& 83.98& 87.01\\
2&93.85&94.48&96.11&\textbf{96.50} & 92.13& 94.89\\
3&70.80&71.48&73.91&\textbf{74.21}&70.85 &73.55\\
5&93.51&93.29&93.12&\textbf{93.68} &90.13 & 92.96\\
6&66.85&68.22&56.36&\textbf{73.07} &70.71 & 67.81\\
7&63.09&64.76&62.69& \textbf{66.59}&63.88 & 64.87\\
8&62.00&62.01&59.89&63.01 &60.17 & \textbf{63.84}\\
12&86.88&86.51&86.92&88.42 & 84.51& \textbf{88.97}\\
13&92.71&92.10&93.06&\textbf{93.55} & 90.01&92.71\\
17&\textbf{93.53}&92.68&91.45&92.27 & 89.43& 92.73\\
18&72.16&70.99&75.41& \textbf{76.78}& 73.25& 74.51\\
19&87.82&87.18&79.76&\textbf{89.88} & 86.01& 88.21\\
21&89.67&88.34&\textbf{92.17}&90.07&85.98 & 89.89\\
22&90.02&90.00&84.93&\textbf{90.48} & 86.60& 90.02\\
24&93.01&93.27&93.27& \textbf{93.77}& 89.39& 93.27\\
25&99.38&99.58&99.43&99.60 &95.19 & \textbf{99.62}\\
33&80.83&89.16&72.82&\textbf{90.91} &87.67 & 89.33\\
34&98.51&98.73&89.66&\textbf{98.79} &94.31 & \textbf{98.79}\\
\midrule
Average & 84.51& 84.94&82.39 &\textbf{86.66} &83.01 & 85.72\\
\bottomrule
     \end{tabular}
\end{adjustbox}
\end{table*}

Tables~\ref{Table:auclin} and~\ref{Table:gmanlin}, respectively, show the AUC and G-mean scores with different imbalanced ratios using a linear kernel function.FSVM-CIL~\cite{Batuwita2010fsvm} and EFTWSVM-CIL~\cite{Gupta2019fuzzy} rank first and second in terms of average AUC and G-mean rates. FSVM-CILFSVM-CIL~\cite{Batuwita2010fsvm} produces AUC and G-mean results of 87.81\% and 87.15\%, respectively. In addition, FSVM-CIL~\cite{Batuwita2010fsvm} and EFTWSVM-CIL~\cite{Gupta2019fuzzy} obtain the best AUC rates for 8 and 7 data sets, respectively, and outperform 10 and 5 data sets in term of G-mean.\par

AUC and G-mean of algorithmic models with a nonlinear kernel function are shown in Tables~\ref{Table:aucnonlin} and~\ref{Table:gmeannon}, respectively. Overall, FSVM-CIL~\cite{Batuwita2010fsvm} outperforms other models in terms of AUC (88.26\%) and G-mean (86.66\%). EFTWSVM-CIL~\cite{Gupta2019fuzzy} is the second best method with AUC and G-man of 87.95\% and 85.72\%, respectively. Both models produce the best AUC rates for 6 data sets. In term of G-mean, FSVM-CIL~\cite{Batuwita2010fsvm} outperforms other methods for 12 data sets.\par

\subsection{Performance comparison between re-sampling and fusion methods}
\label{Sec:sec:enss}

\begin{table*}[tb!]   
\centering
\caption{\label{Table:aucens} AUC (\%) rates of re-sampling and fusion-based SVM models for imbalanced data sets.}
    \begin{adjustbox}{width=\textwidth}
    \begin{tabular}{l c c c c c c c c}
    \toprule
Data \#&FSVM\cite{lin2002fuzzy}&SVM-SMOTE\cite{Nguyen2011Borderline}&SVM-OSS\cite{zar2001Random}&SVM-RUS\cite{zar2001Random}&SVM\cite{cortes1995support}&EasyEnsemble\cite{liu2008exploratory}&AdaBoost\cite{galar2012areview}&1-NN\cite{galar2012areview}\\
\midrule
2&96.32$\pm$1.44&96.64$\pm$0.86&96.61$\pm$0.58&96.35$\pm$1.32&\textbf{96.81$\pm$0.54}&96.58$\pm$0.81&94.43$\pm$2.01&94.40$\pm$1.12\\
3&71.10$\pm$2.11&72.31$\pm$3.49&70.53$\pm$3.98&72.46$\pm$1.02&70.87$\pm$3.36&\textbf{75.22$\pm$1.68}&72.09$\pm$2.53&65.47$\pm$2.19\\
4&69.54$\pm$2.68&70.18$\pm$2.26&68.55$\pm$2.03&69.68$\pm$3.06&71.83$\pm$2.34&\textbf{73.46$\pm$3.74}&67.39$\pm$3.23&64.42$\pm$2.25\\
5&69.41$\pm$4.42&72.40$\pm$3.88&67.31$\pm$4.90&68.17$\pm$4.96&74.14$\pm$5.72&\textbf{97.64$\pm$1.21}&96.83$\pm$0.85&91.49$\pm$2.14\\
6&61.75$\pm$1.09&65.35$\pm$4.10&68.59$\pm$4.22&68.72$\pm$3.54&65.20$\pm$3.47&\textbf{79.45$\pm$3.59}&67.79$\pm$3.58&59.81$\pm$1.78\\
9&82.44$\pm$2.59&85.56$\pm$2.36&79.28$\pm$3.22&83.68$\pm$2.98&81.46$\pm$2.74&\textbf{99.62$\pm$0.29}&99.53$\pm$0.70&99.01$\pm$1.35\\
10&90.08$\pm$2.86&92.47$\pm$1.54&89.65$\pm$2.06&90.12$\pm$1.94&89.29$\pm$5.16&\textbf{93.03$\pm$3.14}&87.48$\pm$2.18&80.59$\pm$2.22\\
11&59.33$\pm$5.96&65.71$\pm$10.82&62.01$\pm$4.53&65.64$\pm$4.26&61.93$\pm$9.72&\textbf{96.16$\pm$1.01}&89.63$\pm$0.86&88.49$\pm$2.39\\
12&87.31$\pm$3.84&88.98$\pm$1.14&89.81$\pm$3.46&88.81$\pm$3.32&87.48$\pm$3.69&\textbf{93.15$\pm$2.01}&84.49$\pm$4.01&86.11$\pm$5.41\\
13&91.32$\pm$10.21&91.48$\pm$9.14&92.16$\pm$9.43&\textbf{95.68$\pm$4.81}&92.21$\pm$11.91&91.59$\pm$8.64&91.38$\pm$10.19&87.35$\pm$11.36\\
14&68.86$\pm$6.24&73.71$\pm$4.01&66.44$\pm$5.49&72.38$\pm$3.89&68.42$\pm$5.08&\textbf{75.64$\pm$4.52}&66.29$\pm$4.15&68.42$\pm$4.99\\
15&78.12$\pm$6.84&\textbf{81.44$\pm$4.83}&81.23$\pm$4.01&79.44$\pm$4.39&80.38$\pm$4.34&80.47$\pm$4.58&72.40$\pm$3.14&77.12$\pm$2.76\\
16&\textbf{93.42$\pm$9.81}&92.18$\pm$9.13&91.42$\pm$10.33&90.73$\pm$8.01&93.21$\pm$6.86&89.56$\pm$5.87&87.68$\pm$12.80&87.68$\pm$12.57\\
17&88.31$\pm$6.35&\textbf{89.12$\pm$7.98}&88.26$\pm$10.19&88.34$\pm$7.29&88.46$\pm$8.82&87.84$\pm$6.72&80.16$\pm$10.03&80.48$\pm$12.38\\
18&80.48$\pm$6.47&81.73$\pm$4.42&81.56$\pm$5.85&\textbf{82.27$\pm$4.25}&79.12$\pm$6.10&80.20$\pm$5.24&68.65$\pm$5.82&71.34$\pm$6.16\\
19&86.43$\pm$3.10&90.39$\pm$3.45&87.90$\pm$5.21&89.36$\pm$3.45&91.45$\pm$2.23&96.72$\pm$2.54&94.68$\pm$4.86&\textbf{100.00$\pm$0.00}\\
20&\textbf{90.14$\pm$6.40}&90.01$\pm$5.55&90.01$\pm$6.48&88.68$\pm$5.35&88.52$\pm$4.89&86.14$\pm$5.68&77.81$\pm$4.33&85.04$\pm$4.76\\
21&91.32$\pm$3.49&\textbf{91.46$\pm$4.84}&88.76$\pm$6.46&89.91$\pm$4.22&90.26$\pm$2.48&87.98$\pm$7.78&87.15$\pm$6.49&62.79$\pm$7.56\\
22&92.27$\pm$6.14&92.27$\pm$6.01&\textbf{93.14$\pm$7.12}&92.44$\pm$4.37&91.49$\pm$5.98&87.70$\pm$10.84&85.12$\pm$13.45&88.43$\pm$12.11\\
23&89.42$\pm$3.01&\textbf{91.79$\pm$3.76}&90.04$\pm$5.48&90.11$\pm$3.20&88.72$\pm$7.96&90.34$\pm$2.28&88.37$\pm$9.49&88.43$\pm$7.51\\
24&91.50$\pm$9.63&91.64$\pm$9.99&\textbf{92.68$\pm$4.48}&90.69$\pm$9.83&90.69$\pm$9.96&89.76$\pm$6.62&79.01$\pm$14.18&87.48$\pm$10.33\\
25&99.75$\pm$0.17&99.75$\pm$0.19&99.75$\pm$0.18&99.75$\pm$0.19&99.70$\pm$0.18&\textbf{99.82$\pm$0.15}&51.06$\pm$0.01&99.68$\pm$1.01\\
26&88.11$\pm$10.40&91.22$\pm$9.84&88.94$\pm$11.68&92.18$\pm$3.49&\textbf{95.98$\pm$1.32}&86.32$\pm$11.48&87.47$\pm$10.12&93.89$\pm$8.87\\
27&93.45$\pm$5.32&94.71$\pm$4.89&93.84$\pm$6.21&94.64$\pm$4.13&\textbf{95.44$\pm$2.01}&90.56$\pm$5.03&84.39$\pm$13.70&88.78$\pm$4.39\\
28&58.36$\pm$3.78&\textbf{67.42$\pm$6.92}&63.38$\pm$4.50&65.72$\pm$7.14&61.64$\pm$18.69&65.59$\pm$5.12&52.13$\pm$0.16&57.18$\pm$5.75\\
29&88.68$\pm$8.26&\textbf{92.45$\pm$8.78}&81.25$\pm$11.65&89.35$\pm$2.99&79.18$\pm$13.42&87.43$\pm$10.19&80.41$\pm$18.64&85.68$\pm$15.34\\
30&77.46$\pm$10.51&78.56$\pm$10.19&77.24$\pm$7.14&79.28$\pm$8.09&80.71$\pm$12.61&\textbf{81.34$\pm$9.75}&75.44$\pm$7.78&75.36$\pm$7.60\\
31&80.68$\pm$5.09&\textbf{85.12$\pm$0.98}&82.53$\pm$3.22&84.58$\pm$2.73&81.76$\pm$4.93&82.61$\pm$3.16&59.40$\pm$4.88&67.59$\pm$6.69\\
32&65.83$\pm$8.39&73.68$\pm$5.15&66.45$\pm$6.89&\textbf{75.71$\pm$4.21}&63.51$\pm$19.80&71.98$\pm$10.88&60.37$\pm$7.76&56.49$\pm$4.16\\
33&95.76$\pm$1.25&\textbf{96.52$\pm$0.67}&95.76$\pm$1.21&96.27$\pm$1.32&95.96$\pm$1.01&95.91$\pm$1.39&86.42$\pm$6.58&84.53$\pm$2.98\\
34&89.51$\pm$17.30&\textbf{89.56$\pm$17.41}&85.78$\pm$15.54&85.40$\pm$16.11&87.44$\pm$8.39&77.43$\pm$14.60&65.79$\pm$17.19&85.59$\pm$15.48\\
35&88.45$\pm$7.61&\textbf{91.34$\pm$4.82}&87.20$\pm$5.75&89.48$\pm$5.32&89.34$\pm$6.59&85.28$\pm$4.71&73.66$\pm$10.12&78.84$\pm$8.82\\
36&55.49$\pm$9.99&66.13$\pm$7.28&54.38$\pm$8.01&67.36$\pm$6.15&50.45$\pm$19.14&\textbf{71.46$\pm$12.89}&50.51$\pm$0.95&52.19$\pm$3.15\\
\midrule
Average & 82.16&84.64 &82.19 & 84.04& 82.53& \textbf{86.18}&77.74 &80\\

\bottomrule
     \end{tabular}
    \end{adjustbox}
\end{table*}

In this experiment, we compare the performance of eight different re-sampling and fusion methods, i.e., FSVM~\cite{lin2002fuzzy}, SVM-SMOTE~\cite{Nguyen2011Borderline}, SVM-OSS~\cite{zar2001Random}, SVM-RUS~\cite{zar2001Random}, SVM~\cite{liu2008exploratory}, EasyEnsemble~\cite{liu2008exploratory}, AdaBoost~\cite{galar2012areview} and 1-NN~\cite{galar2012areview}, are compared.
Note that SVM~\cite{liu2008exploratory}, FSVM~\cite{lin2002fuzzy} and 1-NN~\cite{galar2012areview} are proposed for tackling balanced classification problems. 
All SVMs use linear kernel functions. 
Tables~\ref{Table:aucens} and~\ref{Table:gmeans} present the AUC and G-mean scores along with their standard deviation (SD). Overall, EasyEnsemble~\cite{liu2008exploratory} and SVM-SMOTE~\cite{Nguyen2011Borderline} outperform other methods in terms of average AUC (86.18\%) and G-mean (85.22\%), respectively.\par

\begin{table}[bt!]   
\centering
\caption{\label{Table:gmeans}  G-MEAN (\%) rates of re-sampling and fusion-based SVM models for imbalanced data sets. } 
   \begin{adjustbox}{width=\textwidth}
    \begin{tabular}{l c c c c c c c c c}
    \toprule
Data set&FSVM\cite{lin2002fuzzy}&SVM-SMOTE\cite{liu2008exploratory}&SVM-OSS\cite{liu2008exploratory}&SVM-RUS\cite{liu2008exploratory}&SVM\cite{liu2008exploratory}&EasyEnsemble\cite{liu2008exploratory}&AdaBoost\cite{galar2012areview}&1-NN\cite{galar2012areview}\\
\midrule
2&\textbf{96.28$\pm$3.86}&94.59$\pm$2.30&94.56$\pm$1.55&94.31$\pm$3.54&94.76$\pm$1.45&94.53$\pm$2.17&92.53$\pm$5.39&92.30$\pm$3.00\\

3&71.23$\pm$3.33&72.44$\pm$5.51&70.66$\pm$6.28&72.59$\pm$1.61&71.00$\pm$5.30&\textbf{74.36$\pm$2.65}&72.22$\pm$3.99&65.59$\pm$3.46\\

4&69.51$\pm$3.44&70.15$\pm$2.90&68.52$\pm$2.61&69.65$\pm$3.93&71.80$\pm$3.00&\textbf{73.40$\pm$4.80}&67.36$\pm$4.15&64.39$\pm$2.89\\

5&87.42$\pm$1.74&89.18$\pm$1.53&82.77$\pm$1.93&83.86$\pm$1.95&91.38$\pm$2.25&\textbf{95.20$\pm$0.48}&94.82$\pm$0.33&92.23$\pm$0.84\\

6&67.58$\pm$3.82&71.52$\pm$6.37&72.06$\pm$6.78&73.21$\pm$5.40&69.35$\pm$5.16&\textbf{75.32$\pm$5.58}&72.19$\pm$4.82&63.46$\pm$2.29\\

9&94.59$\pm$0.81&95.17$\pm$0.74&87.96$\pm$1.01&93.11$\pm$0.93&90.47$\pm$0.86&\textbf{99.11$\pm$0.09}&99.08$\pm$0.22&99.01$\pm$0.42\\

10&87.51$\pm$3.14&89.83$\pm$1.69&87.09$\pm$2.26&87.55$\pm$2.13&86.74$\pm$5.67&\textbf{90.38$\pm$3.45}&84.98$\pm$2.39&78.29$\pm$2.44\\

11&79.98$\pm$3.95&78.24$\pm$7.17&73.28$\pm$3.00&78.15$\pm$2.82&73.18$\pm$6.44&\textbf{93.54$\pm$0.67}&90.37$\pm$0.57&89.84$\pm$1.58\\

12&85.73$\pm$8.19&87.37$\pm$2.43&88.18$\pm$7.38&87.20$\pm$7.08&84.37$\pm$7.87&\textbf{91.46$\pm$4.29}&82.96$\pm$8.55&84.55$\pm$11.54\\

13&95.36$\pm$4.51&95.53$\pm$4.04&96.24$\pm$4.16&\textbf{97.93$\pm$2.12}&96.29$\pm$5.26&95.64$\pm$4.26&95.42$\pm$5.03&91.21$\pm$5.02\\

14&68.01$\pm$1.42&72.80$\pm$0.91&65.62$\pm$1.25&71.49$\pm$0.88&67.57$\pm$1.16&\textbf{74.71$\pm$1.03}&65.47$\pm$0.94&67.57$\pm$1.13\\

15&76.48$\pm$8.80&\textbf{79.70$\pm$6.21}&79.49$\pm$5.16&77.74$\pm$5.65&75.45$\pm$5.58&78.75$\pm$5.89&70.85$\pm$4.04&75.47$\pm$3.55\\

16&\textbf{89.32$\pm$8.57}&88.13$\pm$7.97&87.41$\pm$9.02&86.75$\pm$7.00&89.12$\pm$5.99&85.63$\pm$5.13&83.83$\pm$11.18&83.80$\pm$10.98\\

17&91.90$\pm$8.88&\textbf{92.74$\pm$11.16}&91.85$\pm$12.25&91.93$\pm$10.19&92.06$\pm$12.33&91.41$\pm$9.40&83.42$\pm$9.54&83.75$\pm$10.12\\

18&78.39$\pm$8.09&79.61$\pm$5.53&79.44$\pm$7.31&\textbf{80.13$\pm$5.31}&77.06$\pm$7.63&78.12$\pm$6.55&66.87$\pm$7.28&69.49$\pm$7.70\\

19&86.28$\pm$6.02&90.23$\pm$6.70&87.75$\pm$10.12&89.20$\pm$6.70&91.29$\pm$4.33&95.55$\pm$4.39&93.51$\pm$9.44&\textbf{99.83$\pm$0.20}\\

20&\textbf{88.13$\pm$6.40}&88.00$\pm$5.50&87.89$\pm$6.44&86.70$\pm$5.31&86.55$\pm$4.86&84.22$\pm$5.64&76.07$\pm$4.30&83.14$\pm$4.73\\

21&87.58$\pm$8.62&\textbf{87.71$\pm$11.95}&85.12$\pm$12.95&86.23$\pm$10.42&86.56$\pm$6.12&84.38$\pm$13.21&83.58$\pm$11.02&60.22$\pm$13.67\\

22&90.48$\pm$3.95&\textbf{92.27$\pm$3.87}&90.41$\pm$4.58&90.65$\pm$2.81&89.71$\pm$3.85&86.00$\pm$6.97&83.47$\pm$8.65&86.71$\pm$7.79\\

23&92.84$\pm$2.03&\textbf{95.30$\pm$2.54}&93.48$\pm$3.70&93.19$\pm$2.16&92.11$\pm$5.37&93.79$\pm$1.54&91.75$\pm$6.40&91.82$\pm$5.06\\

24&91.90$\pm$1.48&92.04$\pm$1.53&\textbf{93.08$\pm$0.69}&91.00$\pm$1.51&91.09$\pm$1.53&90.15$\pm$1.02&79.35$\pm$2.18&87.86$\pm$1.59\\

25&99.86$\pm$0.11&99.89$\pm$0.05&99.86$\pm$0.7&99.87$\pm$0.05&99.81$\pm$0.07&\textbf{99.93$\pm$0.04}&81.09$\pm$0.41&99.79$\pm$0.65\\

26&85.58$\pm$9.96&88.60$\pm$9.42&86.39$\pm$11.18&89.53$\pm$3.34&\textbf{90.84$\pm$2.22}&83.84$\pm$10.99&84.96$\pm$9.69&90.19$\pm$8.49\\

27&93.05$\pm$5.11&94.30$\pm$4.70&93.44$\pm$5.96&94.23$\pm$3.97&\textbf{94.43$\pm$1.93}&90.17$\pm$4.83&84.03$\pm$13.16&88.40$\pm$4.22\\

28&61.30$\pm$6.81&\textbf{70.82$\pm$12.57}&66.57$\pm$8.11&69.03$\pm$12.86&64.74$\pm$15.66&68.89$\pm$9.22&54.76$\pm$0.29&60.06$\pm$10.36\\

29&73.43$\pm$10.28&\textbf{76.55$\pm$10.93}&67.28$\pm$14.50&73.98$\pm$3.72&65.56$\pm$16.70&72.39$\pm$12.68&66.58$\pm$13.19&70.95$\pm$13.09\\

30&71.94$\pm$6.61&72.96$\pm$6.41&70.73$\pm$4.49&73.63$\pm$5.09&74.96$\pm$7.93&\textbf{75.54$\pm$6.13}&70.06$\pm$4.89&69.99$\pm$4.78\\

31&80.80$\pm$5.76&\textbf{85.25$\pm$1.11}&82.65$\pm$3.64&84.70$\pm$3.09&81.88$\pm$5.58&82.73$\pm$3.58&59.49$\pm$5.52&67.69$\pm$7.57\\

32&66.22$\pm$4.54&74.11$\pm$4.54&66.84$\pm$6.07&\textbf{74.61$\pm$3.71}&63.89$\pm$11.44&72.36$\pm$9.58&60.73$\pm$6.83&56.82$\pm$3.66\\

33&90.42$\pm$2.21&\textbf{91.14$\pm$1.18}&90.38$\pm$2.14&90.91$\pm$2.33&90.61$\pm$1.79&90.56$\pm$2.46&81.60$\pm$11.63&79.82$\pm$5.27\\

34&98.24$\pm$3.42&\textbf{98.29$\pm$3.44}&94.15$\pm$3.07&93.73$\pm$3.18&95.97$\pm$1.66&84.98$\pm$2.89&72.21$\pm$3.40&93.94$\pm$3.06\\

35&89.57$\pm$4.20&\textbf{92.50$\pm$2.66}&88.30$\pm$3.17&90.61$\pm$2.94&90.47$\pm$3.64&86.36$\pm$2.60&74.59$\pm$5.59&79.83$\pm$4.87\\

36&54.74$\pm$2.54&65.25$\pm$1.85&53.64$\pm$2.04&66.45$\pm$1.56&49.77$\pm$4.87&\textbf{70.49$\pm$3.28}&49.83$\pm$0.24&51.48$\pm$0.80\\
\midrule
Average &83.08 & \textbf{85.22}&82.52 &84.36 & 82.75& 84.97&77.88 &79.38\\

\bottomrule
     \end{tabular}
    \end{adjustbox}
\end{table}

\section{Discussion}
\label{Sec:dis}
This section analyzes and discusses the main findings of this review and highlights several future research directions along with several research questions.

SVM and its variants for learning imbalanced data can be broadly categorized into three: re-sampling, algorithmic, and fusion methods.  Re-sampling methods pre-process data samples by attempting to balance between the numbers of majority and minority class samples. This can be achieved by under-sampling the majority class samples, over-sampling the minority class samples, or their combination.  Despite the success of re-sampling methods, several limitations exist.  These methods require pre-processing of data samples before model training. Although under-sampling strategies can reduce the computational burden, useful information could be eliminated by removing data samples close to the decision boundary.  On the other hand, over-sampling strategies could lead to the overfitting problem, and they require longer execution durations.

Comparing with re-sampling methods, algorithmic models do not require data pre-processing, therefore a lower computational burden.  In general, two algorithmic strategies that enable the SVM and its variants to effectively deal with imbalanced data are: (i) applying different cost functions to the majority and minority class samples; (ii) modifying the underlying kernel functions. On the other hand, fusion models combine re-sampling and algorithmic methods or employ an ensemble of learning models, usually weak learners, to tackle class-imbalanced learning problems.  As reported in the literature, fusion models usually outperform algorithmic methods. However, they are computationally expensive, as it is necessary to train multiple models in fusion methods.  Table 8 presents a comparison of the three categories of SVM-based models from different aspects. Table~\ref{Table:com} presents a comparison of the three categories of SVM-based models from different aspects.
The following sub-section presents several directions that need further investigation.

\begin{table}[h] 
\centering
\caption{\label{Table:com}Comparative analysis of SVM-based techniques for imbalanced class learning.}
   \begin{adjustbox}{width=0.99\textwidth}
   
    \begin{tabular}{p{0.15\textwidth} p{0.3\textwidth} p{0.3\textwidth} p{0.3\textwidth}}
    \toprule  
\textbf{Aspect} & \textbf{Re-sampling Methods} & \textbf{Algorithmic Methods} & \textbf{Fusion Methods} \\ \midrule  
\textbf{Description} & Pre-process data samples to balance the numbers of majority and minority class samples. & Modify the underlying algorithm of SVM and its variants directly without data pre-processing. & Integrate two or more methods into a single framework. \\ \midrule  
\textbf{Techniques} & Under-sampling or over-sampling, or a combination of both. & Use different cost functions. Modify kernel functions. & Combine re-sampling and algorithmic techniques. Ensemble of learning models. \\ \midrule  
\textbf{Advantages} & Reduce computational burden (under-sampling). Enhance minority class representation (over-sampling). & Possess a lower computational load. Direct modification of SVM algorithms. & Generally perform better than algorithmic methods alone. Address both data and algorithmic issues. \\ \midrule  
\textbf{Disadvantages} & Risk of information loss (under-sampling). Risk of overfitting (over-sampling). Require data pre-processing. & Potentially less effective in handling highly imbalanced data. & Computationally expensive owing to the need to train multiple models. \\ \midrule  
\textbf{Execution Time} & Vary subject to the extent of re-sampling. & Generally faster as no data pre-processing is needed. & Time-consuming owing to the complexity and multiple model training. \\

     \bottomrule
     \end{tabular}
   \end{adjustbox}
\end{table}

\subsection{Future research directions}
\label{Sec:sec:research}
Despite the success of SVM and its variants in solving imbalanced data learning problems, there are a number of challenging issues pertaining to the availability of sufficient minority class samples for training purposes. Several future research directions that require further investigations are:

\begin{itemize}
    \item \textbf{Noisy data sets:} SVM and its variants are highly sensitive to noisy samples, which becomes more challenging in tackling imbalanced data. Although several studies have discussed this issue, comprehensive theoretical and empirical analyses are still lacking.
    \item \textbf{Large scale data sets:} The advent of big data has resulted in various tasks with large-scale and low-quality data.  When large data samples are available for training, the underlying learning methods should be adaptive toward both space and time requirements. This challenge becomes more significant when dealing with imbalanced data. Like many other machine learning algorithms, SVM and its variants can be improved when they are applied to processing large-scale imbalanced data sets.
    
    \item \textbf{Extremely imbalanced data sets:} In big data, identifying the minority class samples becomes more difficult due to bias towards the majority class samples. Such a biased learning process can result in recognizing all instances as the majority class samples and producing deceptively high accuracy rates. Many studies in the current literature lack in-depth investigations on both theoretical and practical aspects of SVM and its variants, making it difficult to conclude whether one method is more effective and efficient than another in tackling extremely imbalanced data.
\end{itemize}

Based on the above analysis, a number of research questions that are of interest to both researchers and practitioners are as follows:
\begin{itemize}
\item What is the impact of class noise on classification algorithms trained on highly skewed data sets?
\item Are the effects of class noise in imbalanced data sets uniform across different classification alhorithms?
\item How do sampling procedures, which are often used to alleviate class imbalanced issues, perform in the presence of class noise?
\item How do different sampling techniques work with respect to different levels of noise in different proportions of both majority and minority classes?
\end{itemize}

\subsection{Concluding remarks}
\label{Sec:sec:con}
Enabling SVM to learn from class imbalanced data is useful for handling many real-world applications, such as medical diagnosis or fault detection, in which obtaining sufficient samples for all classes is impractical. In this study, a review of SVM-based models for class imbalanced learning has been presented. Specifically, we have categorized the existing methods into three: re-sampling, algorithmic and fusion methods. We have also discussed the principles of several representative models of each category, and conducted an empirical study to compare the relevant methods. The outcome indicates that fusion methods usually perform better than other methods.  Nonetheless, they are computationally expensive, as they require the use of multiple classifiers.  In addition, several directions for future research have been discussed, and our reviews provide the necessary guidelines for researchers from different fields to utilize SVM and its variants for effectively learning from class-imbalanced data sets.\\

\textbf{Declaration of competing interest}:\\
The corresponding author certifies that all authors have read the manuscript before submission, and there is no conflict of interest associated with this publication, and there has been no financial support for this work that could have influenced its outcome.\\

\textbf{Data availability statement}:\\
\\
The data that support the findings of this study are openly available in UCI and KEEL repositories at http:// archive.ics.uci.edu/ml/datasets.html and https://sci2s.ugr.es/keel/imbalanced.php, respectively.\\

\textbf{Credit authorship contribution statement}:\\
\\
\textbf{Salim Rezvani}:  Data curation and validation. \textbf{Farhad Pourpanah}: Conceptualization, Investigation, Writing –original draft, review \& editing, Supervision. \textbf{Chee Peng Lim}: Writing – review \& editing, Supervision. \textbf{Q. M. Jonathan Wu}: Funding acquisition, Supervision.

==========
\bibstyle{sn-apacite}
\bibliography{sn-bibliography}

\end{document}